\documentclass[journal]{IEEEtran}
\usepackage{amsmath,graphicx,bm,booktabs,multirow}
\usepackage{threeparttable}
\usepackage{algorithm}
\usepackage{algorithmic}
\usepackage{subfig}
\usepackage{xcolor}
\usepackage{rotating}
\usepackage{longtable}
\usepackage{lscape}     
\usepackage{cite}
\newtheorem{remark}{\textbf{Remark}}

\ifCLASSINFOpdf
\else
\fi
\begin{document}
%
\title{Few-Shot Defect Segmentation Leveraging Abundant Normal Training Samples Through Normal Background Regularization and Crop-and-Paste Operation}

\author{Dongyun~Lin,~\IEEEmembership{Member,~IEEE,}
        Yanpeng~Cao,~\IEEEmembership{Member,~IEEE,}
        Wenbing~Zhu,%
        ~and~Yiqun Li
\thanks{Dongyun Lin and Yiqun Li are with Institute for Infocomm Research (I2R), the Agency for
Science, Technology and Research (A*STAR), Singapore}
\thanks{Yanpeng Cao and Wenbing Zhu are with State Key Laboratory of Fluid Power and Mechatronic Systems, College of Mechanical Engineering and Key Laboratory of Advanced Manufacturing Technology of Zhejiang Province, College of Mechanical Engineering, Zhejiang University, Hangzhou, 310027, China}}

\maketitle


\begin{abstract}
In industrial product quality assessment, it is essential to determine whether a product is defect-free and further analyze the severity of anomality. To this end, accurate defect segmentation on images of products provides an important functionality. In industrial inspection tasks, it is common to capture abundant defect-free image samples but very limited anomalous ones.  Therefore, it is critical to develop automatic and accurate defect segmentation systems using only a small number of annotated anomalous training images. This paper tackles the challenging few-shot defect segmentation task with sufficient normal (defect-free) training images but very few anomalous ones. We present two effective regularization techniques via incorporating abundant defect-free images into the training of a UNet-like encoder-decoder defect segmentation network. We first propose a Normal Background Regularization (NBR) loss which is jointly minimized with the segmentation loss, enhancing the encoder network to produce distinctive representations for normal regions. Secondly, we crop/paste defective regions to the randomly selected normal images for data augmentation and propose a weighted binary cross-entropy loss to enhance the training by emphasizing more realistic crop-and-pasted augmented images based on feature-level similarity comparison. Both techniques are implemented on an encoder-decoder segmentation network backboned by ResNet-34 for few-shot defect segmentation. Extensive experiments are conducted on the recently released MVTec Anomaly Detection dataset with high-resolution industrial images. Under both 1-shot and 5-shot defect segmentation settings, the proposed method significantly outperforms several benchmarking methods.


\end{abstract}


%
\IEEEpeerreviewmaketitle

\section{Introduction}
\label{sec:intro}

\IEEEPARstart{D}{efect} segmentation provides an important step in various industrial inspection tasks relying on localizing and delineating defective regions that visually appear anomalous against the normal ones. In general, it is tedious and time-consuming to manually identify and annotate the defect regions on real-captured images. Therefore, automating the process is in demand. Moreover, in real-world applications, it is relatively easy to capture defect-free images but challenging to obtain sufficient anomalous training images~\cite{wang2018simple,razavi2017integrated}. Hence, building up an automatic and accurate defect segmentation model using only a small number of annotated anomalous images is valuable. Motivated by the aforementioned demands, this paper tackles the task of defect segmentation with very few annotated anomalous images, i.e., few-shot defect segmentation.


Recently, deep neural networks have been widely exploited in many vision based tasks~\cite{ronneberger2015u,long2015fully,ye2021deep,ye2018hierarchical}. Defect segmentation is closely related to the task of Anomaly Detection (AD), which aims at predicting an image (as a whole) to be either normal or anomalous~\cite{pimentel2014review}. Most research works handle AD by exploiting deep neural networks, e.g., Deep Convolution Autoencoder (DCAE) or Generative Adversarial Network (GAN), which are trained using only normal images~\cite{chen2017outlier,ruff2018deep,perera2019ocgan,schlegl2017unsupervised,zhou2019attention}. Although the AD methods achieve good accuracy on classifying low resolution images, they are rarely formulated to accurately localize the defect regions from high-resolution industrial images. Hence, defect segmentation presents a more challenging task which not only requires differentiating anomalous/normal image patches but also delineating the boundaries of the defects within the anomalous image.

Another related task with defect segmentation is semantic segmentation, i.e., predicting the semantic class for every pixel within an image. Semantic segmentation tasks are predominantly tackled using fully convolutional networks (FCN) \cite{long2015fully} which exploit encoder-decoder architectures to predict masks from images ~\cite{chen2017deeplab,ronneberger2015u,iglovikov2018ternausnet,shvets2018angiodysplasia,shvets2018automatic}. Supervised training of these models typically requires a large number of $\left \langle \rm{image}, \rm{mask} \right \rangle$ pairs for tuning network parameters. However, in real-world industrial applications, only very limited anomalous images with pixel-level annotation masks are available for model training. Hence, the traditional supervised training severely suffers from the overfitting problem. Recently, learning with less training data has drawn great research interest in the computer vision community, especially in the field of image classification and segmentation, i.e., few-shot image classification~\cite{vinyals2016matching,sung2018learning,gidaris2018dynamic} and segmentation~\cite{zhang2019canet,wang2019panet,nguyen2019feature}. The state-of-the-art approaches to the few-shot problems typically adopt the meta-learning scheme, which requires an auxiliary dataset with sufficient data to train the meta learner with few-shot learning capability~\cite{vinyals2016matching,sung2018learning,gidaris2018dynamic,zhang2019canet}. On one hand, differing from the generic few-shot semantic segmentation task, sufficient defect-free training images are available in industrial inspection tasks, which are explored to improve the performance of defect segmentation under few-shot settings in this paper.  On another hand, our solution does not require auxiliary datasets for meta-learning as they are usually not available in industrial inspection applications.

To address few-shot defect segmentation from high resolution images, we propose two regularization methods by leveraging abundant defect-free normal images through a novel training scheme involving two input branches during the training of the segmentation network. One is for sufficient defect-free normal images while the other is for limited annotated defective images. The proposed techniques could be easily implemented in an encoder-decoder segmentation networks to significantly improving the accuracy of defect segmentation results under both 1-shot and 5-shot settings. To our best knowledge, this paper is the first work to develop more effective defect segmentation via harvesting of abundant normal training samples under few-shot settings. The major contributions of this paper are summarized as follows:




\begin{enumerate}
   \item We propose a novel Normal Background Regularization (NBR) loss into the training process of the encoder network.  This loss facilitates the encoder to produce distinctive representations of normal regions by maximizing the similarity between the normal regions within anomalous images and randomly picked normal training images.
   \item We propose a simple yet effective Crop-and-Paste (CaP) operation for data augmentation. Moreover, we develop a weighted binary cross-entropy loss to enhance the influence of more realistic defective images via computing the feature-level similarity between artificially generated and real-captured anomalous images.
   \item We adopt a UNet-like encoder-decoder segmentation network backboned by ResNet-34 as the baseline model and demonstrate the effectiveness of the proposed regularization techniques, achieving significantly higher few-shot segmentation and anomaly detection accuracy without increasing inference computational complexity on the MVTec Anomaly Detection benchmark dataset.
\end{enumerate}

The remaining sections of this paper are organized as follows. The related works are reviewed in Section \ref{RelatedWork}. The details of the proposed method are presented in Section \ref{sec:method}. The experimental results on the comparison with several benchmarking methods and the ablation studies are shown and analyzed in Section \ref{sec:experiments}. Section \ref{sec:conclusion} concludes the entire paper.

\section{Related Work}
\label{RelatedWork}
In this section, we briefly review some related works on anomaly detection, U-Net segmentation architecture and few-shot semantic segmentation via meta learning.

\subsection{Anomaly Detection}
\label{AnomalyDetection}
Anomaly detection (AD) aims at predicting a testing image as either normal or anomalous given only normal images as training data. Several early statistical AD methods, such as One-Class Support Vector Machine (OC-SVM)~\cite{scholkopf2001estimating}, One-Class Support Vector Data Description (OC-SVDD)~\cite{Tax2004} and Kernel Density Estimation (KDE)~\cite{10.2307/2237880} address AD using hand-crafted features for one-class classification. Recently, deep learning based methods are widely exploited for anomaly detection problems. These methods can be divided into two groups: image reconstruction based and AD oriented objective based methods. The image reconstruction based methods exploit only normal images to train an image reconstruction model, e.g., Deep Convolution Autoencoder (DCAE) and Deep Generative Adversarial Network (GAN), and assume the model could produce lower reconstruction errors for normal images than those anomalous ones~\cite{chen2017outlier,amarbayasgalan2018unsupervised,sakurada2014anomaly,schlegl2017unsupervised}. The reconstruction based AD methods are built upon the assumption that the reconstruction process is discriminative and they do not directly optimize for an AD oriented objective\cite{ruff2018deep}. To better align the problem target and the optimization objective, AD oriented objective based methods are proposed. For example, Deep SVDD~\cite{ruff2018deep} trains a deep neural network by minimizing the volume of a hypersphere that encloses the latent representations of normal class images.


Our few-shot defect segmentation task is not the traditionally defined AD task. They are similar since they both adopt a sufficient number of normal training images. However, they are different in two aspects: (i) our task not only focus on predicting the whole image as normal or anomalous, but also segmenting out the defective regions which are not provided by AD methods. Towards this more challenging objective, we extend training resources of AD to incorporate very few (1 or 5) annotated anomalous samples; (ii) our task can handle high-resolution industrial images which are more challenging than low resolution images handled by AD methods.

\subsection{U-Net Segmentation Architecture}
\label{U_Net}
In this subsection, we review several U-Net segmentation architectures since we exploit U-Net-like segmentation model in this work. U-Net~\cite{ronneberger2015u} is one of the most widely studied fully convolutional network (FCN) \cite{long2015fully} models which exploit encoder-decoder architectures to predict segmentation masks from images. U-Net consists of an encoding path followed by an decoding path to perform dense pixel-wise prediction. It also introduces skip connections to concatenate low-level features in the encoding path with the high-level features in the decoding path for recovering spatial information. Motivated by the original U-Net, many U-Net variants are proposed to modify the original model from various perspectives~\cite{iglovikov2018ternausnet,shvets2018automatic,chaurasia2017linknet,zhou2019unet++,oktay2018attention,zhang2018road}. TernausNet-11~\cite{iglovikov2018ternausnet}, TernausNet-16~\cite{shvets2018automatic} and AlbuNet-34~\cite{buslaev2018fully} exploited VGG-11, VGG-16 and ResNet-34 as the backbones of their encoding paths, respectively. Ref \cite{zhou2019unet++} proposed U-Net++ which builds up intermediate layers into the skip connections to alleviate big semantic gaps caused by long skip connections. Attention U-Net~\cite{oktay2018attention} is designed by incorporating attention mechanism into the skip connections to make the model focus on salient features within images. Motivated by ResNet~\cite{he2016deep}, \cite{zhang2018road} proposed deep residual U-Net by adding residual (shortcut) connections in the U-Net decoder.

\subsection{Few-shot Semantic Segmentation via Meta Learning}
\label{Few_Shot}

Learning with an extremely small number of training data has drawn great research interest in the computer vision community, especially in the field of image classification, i.e., few-shot image classification~\cite{vinyals2016matching,sung2018learning,gidaris2018dynamic}. Regarding semantic segmentation of images, the regular supervised training of fully-convolutional network models typically requires a large number of $\left \langle \rm{image}, \rm{mask} \right \rangle$ pairs for tuning network parameters. An increasing number of research works are proposed to perform semantic segmentation under few-shot settings~\cite{zhang2019canet,wang2019panet,nguyen2019feature}. The state-of-the-art approaches to addressing the few-shot semantic segmentation usually adopt a learning scheme called meta-learning which requires an auxiliary dataset with sufficient data to train the model for few-shot learning capability instead of fitting the original training data. For example, CANet~\cite{zhang2019canet} is proposed to perform class-agnostic deep feature comparison. PANet~\cite{wang2019panet} is formulated to generate class prototypes for dense pixelwise prediction.

Our few-shot defect segmentation task is similar to few-shot semantic segmentation in terms of only an extremely small number of annotated $\left \langle \rm{image}, \rm{mask} \right \rangle$ training data are available. However, in this work, we aim to formulate a solution that does not require auxiliary datasets for meta-learning as they are usually not available in industrial inspection applications. Instead of using auxiliary datasets, we leverage abundant defect-free training images to improve the defect segmentation performance under few-shot settings.


\section{Method}
\label{sec:method}

\subsection{Problem Setting}
\label{subsec:problem_setting}
We define our $K$-shot defect segmentation problem corresponding to $C$ classes of defects and each has $K$ $\left \langle \rm{image}, \rm{mask} \right \rangle$ pairs of training data. Specifically, the training set $\mathcal{D}_{train}$ consists of two subsets: a normal training subset $\mathcal{D}^{n}_{train} = {\{\bm{I}^{n}_i\}}$ where $i=1,2,\cdots,N_n$ and a defective training subset of $\left \langle \rm{image}, \rm{mask} \right \rangle$ pairs $\mathcal{D}^{d}_{train}={\{(\bm{I}^{d}_{c,k}, \bm{M}^{d}_{c,k})\}}$ where $n$ denotes ``normal'', $d$ denotes ``defective'', $c(c=1,2,\cdots, C)$ and $k(k=1,2,\cdots,K)$ denote the indexes for the defective categories and the defective images, respectively. The number of normal images $N_{n}$ is sufficiently large while these images contain no mask annotation since they are defect-free. The testing set $\mathcal{D}_{test}$ contains multiple unseen anomalous images with pixel-wise annotated masks from one of $C$ training defect classes. We exploit both normal and defective training subsets to train a defect segmentation network whose performance is evaluated in the testing dataset.

\begin{figure}[!t]
\centering
   \includegraphics[scale=0.45]{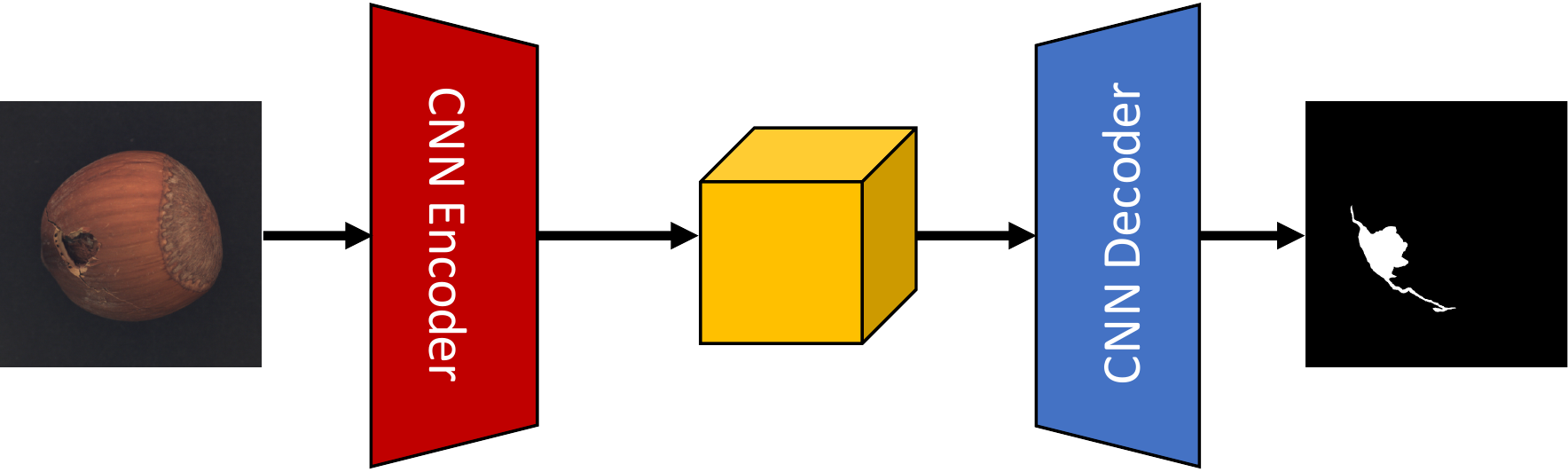}
   \caption{The illustration of the testing process. For clarity, the skip connections for the U-Net model is not plotted.}
\label{fig:TestProcess}
\end{figure}

\begin{figure}[!t]
\centering
   \includegraphics[scale=0.3]{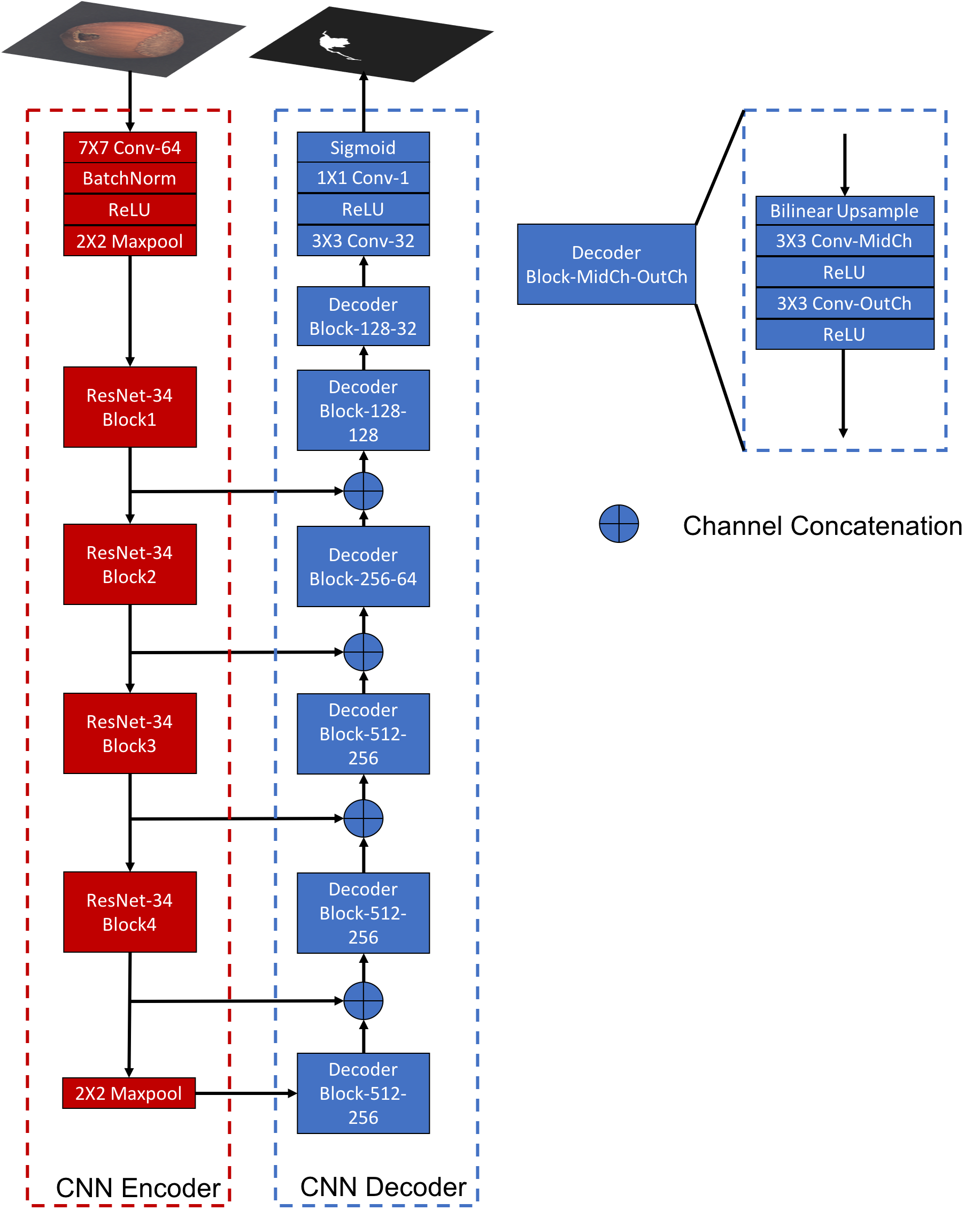}
   \caption{The illustration of the baseline segmentation model~\cite{buslaev2018fully}. $\left \langle \rm{k} \times \rm{k} \right \rangle$ \rm{Conv}-\rm{OutCh} denotes the convolution with kernel size $\rm{k}$ and output channel number $\rm{OutCh}$. $\left \langle \rm{k} \times \rm{k} \right \rangle$ $\rm{Maxpool}$ denotes maxpooling operation with kernel size $\rm{k}$. BatchNorm denotes batch normalization. ReLU and Sigmoid denotes Rectified Linear Unit and sigmoid activation function, respectively. The architecture of each of the ResNet-34 blocks is specified in \cite{he2016deep}.}
\label{fig:Baseline}
\end{figure}

\subsection{Method Overview}
\label{subsec:method_overview}

To alleviate the overfitting problem due to the limited annotated anomalous image samples, two regularization techniques are proposed by exploiting the abundant defect-free images into the training of a segmentation network. The first contribution is the proposal of a  novel normal background regularization (NBR) loss, which is jointly minimized with the segmentation loss to encourage the encoder network to produce distinctive representations of normal regions. The second contribution is to deploy the Crop-and-Paste (CaP) operation to generate artificial defect samples for data augmentation and further design a weighted cross-entropy loss to enhance the influence of more realistic defective images during the training process.  Based on the proposed regularization techniques, a U-Net-like encoder-decoder segmentation network is trained for few-shot defect segmentation. During inference, the unseen images are fed through the segmentation network to predict the defect masks, as illustrated in Fig.\ref{fig:TestProcess}.



\begin{figure*}[htpb]
\centering
   \includegraphics[scale=0.38]{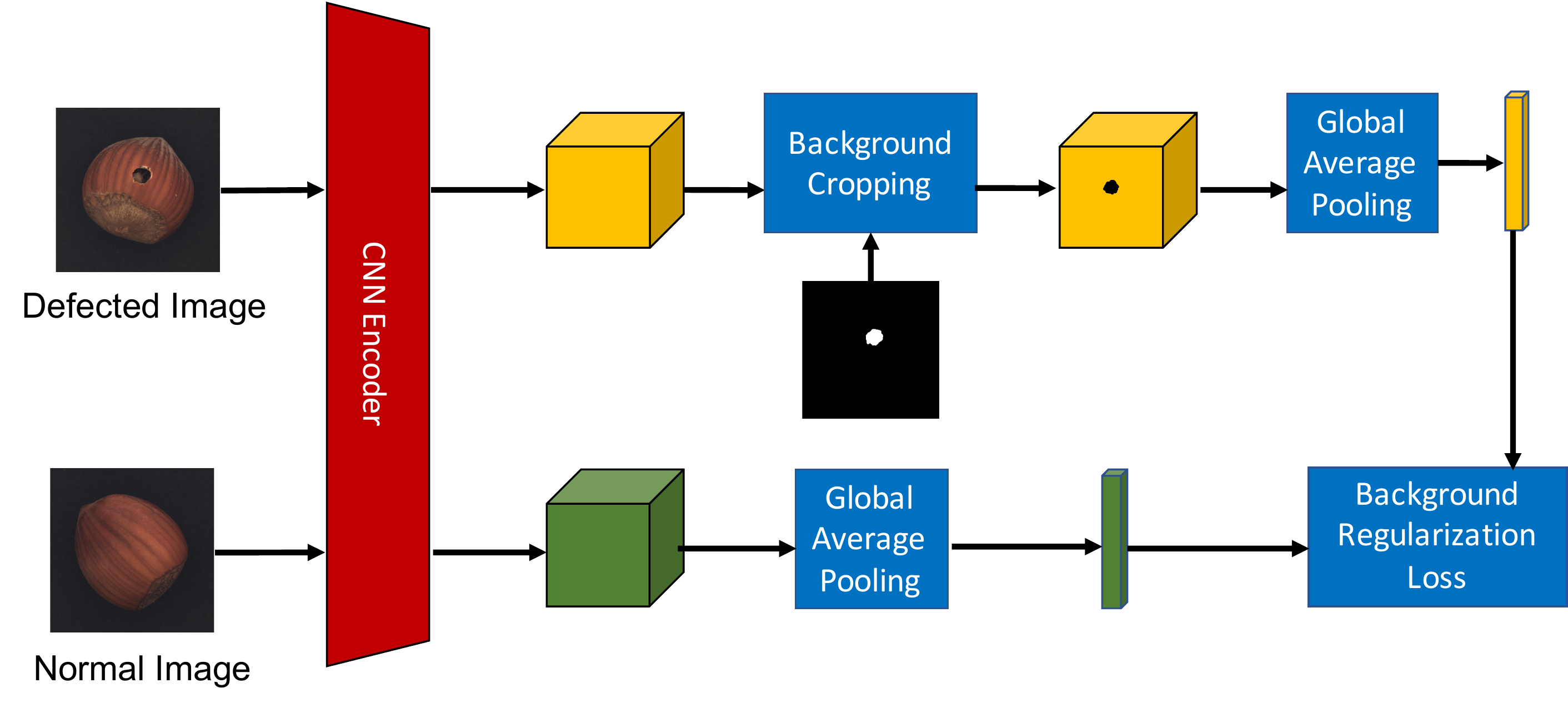}
   \caption{The illustration of Normal Background Regularization.}
\label{fig:NBR}
\end{figure*}

\subsection{Baseline Encoder-Decoder Segmentation Network}
\label{subsec:Baseline}
For baseline network, we adopt a U-Net-like encoder-decoder segmentation network proposed in~\cite{buslaev2018fully} backboned by ResNet-34 shown in Fig.~\ref{fig:Baseline}. The encoder part exploits the convolution blocks of ResNet-34 with fully-connected layers removed. The decoder part exploits several decoder blocks to progressively upsample the feature maps. The skip connections proposed by U-Net~\cite{ronneberger2015u} are applied to concatenate low-level feature maps with high-level ones. An image can be fed through the CNN encoder and decoder network in sequence to obtain the predicted defect mask as shown in Fig.~\ref{fig:Baseline}.


\subsection{Normal Background Regularization (NBR)}
\label{subsec:NBR}
We first propose $\textbf{N}$ormal $\textbf{B}$ackground $\textbf{R}$egularization (NBR) to facilitate the encoder to produce distinctive representations of normal regions by maximizing the similarity between the normal regions within anomalous images and the global regions of normal training images. Such idea is motivated by the assumption that the segmentation network should produce relatively consistent deep representations for the normal regions within anomalous images to the normal training images because they have similar appearances. Given a large number of normal (defect-free) training images but very few annotated anomalous ones, we attempt not only to delineate defective regions but also to maximize the similarity between the normal regions within anomalous images and abundant normal training samples. In such a way, it is possible to better depict the characteristics of normal regions and to generate more accurate boundaries between defective and defect-free regions within an image.

Specifically, as shown in Fig.~\ref{fig:NBR}, assuming batch size is set as 1, i.e., within one minibatch, we have one normal image $\bm{I}^n$, one defect image $\bm{I}^d$ and its defect mask $\bm{M}^d$. Both $\bm{I}^n$ and $\bm{I}^d$ are fed through the encoder to generate feature map $\bm{F}^n$ and $\bm{F}^d$, respectively. Then the normal background component $\bm{B}^d$ is cropped from $\bm{F}^d$ by elementwisely multiplying $\bm{F}^d$ with the reversed downsampled mask $\hat{\bm{M}}^d$:
\begin{equation}
\bm{B}^d = \bm{F}^d \odot (\bm{1} - \hat{\bm{M}}^d),
\label{eq:Bd}
\end{equation}
where $\hat{\bm{M}}^d$ is the downsampled version of the original groundtruth mask $\bm{M}^d$. $\bm{1}$ is the mask with the same size as $\hat{\bm{M}}^d$ filled with ones. $\odot$ denotes elementwise multiplication. Then Global Average Pooling (GAP)~\cite{lin2013network} is applied on both $\bm{B}^d$ and $\bm{F}^n$ to obtain vectorized representations $\bm{b}^d$ and $\bm{f}^n$ as:
\begin{equation}
\bm{b}^d = {\rm{GAP}}(\bm{B}^d),\quad \bm{f}^n = {\rm{GAP}}(\bm{F}^n).
\label{eq:bd}
\end{equation}
To encourage alignment of normal regions, a normal background regularization loss is proposed as the negative cosine similarity between $\bm{b}^d$ and $\bm{f}^n$ as:
\begin{equation}
\mathcal{L}_{\rm{NBR}} = -\frac{(\bm{b}^d)^{\rm{T}}(\bm{f}^n)}{||\bm{b}^d||_2 \cdot||\bm{f}^n||_2}.
\label{eq:LNBR}
\end{equation}
The normal background regularization loss is jointly minimized with the segmentation loss proposed in the subsequent subsection.

\begin{algorithm}[!t]\small
\caption{Training procedure}
\begin{algorithmic}
\label{alg:AGF}
\STATE{\textbf{Input}:
\begin{itemize}
  \item The normal training subset $\mathcal{D}^{n}_{train} = {\{\bm{I}^{n}_i\}}$ ($i=1,2,\cdots,N_n$)
  \item The defect training subset $\mathcal{D}^{d}_{train}={\{(\bm{I}^{d}_{c,k}, \bm{M}^{d}_{c,k})\}}$ ($c=1,2,\cdots,C$ and $k=1,2,\cdots,K$)
  \item The encoder-decoder segmentation network
\end{itemize}}
\FOR{number of iterations on defect batch sampling}

    \STATE{$\bullet$ Sample a minibatch of $m$ training $\left \langle \rm{image}, \rm{mask} \right \rangle$ pairs $\{ (\bm{I}^{d}_{c,1}, \bm{M}^{d}_{c,1}), \dots, (\bm{I}^{d}_{c,m}, \bm{M}^{d}_{c,m}) \}$ from $\mathcal{D}^{d}_{train}$.}
    \STATE{}
    \FOR{number of iterations on normal batch sampling}

    \STATE{$\bullet$ Sample a minibatch of $m$ training normal images $\{ \bm{I}^{n}_{1}, \dots, \bm{I}^{n}_{m} \}$ from $\mathcal{D}^{n}_{train}$.}

    \STATE{}
    \STATE{\textbf{Normal Background Regularization}}
    \STATE{}
    \STATE{$\bullet$ Calculate $\mathcal{L}_{\rm{NBR}}(\bm{I}^{n}_{i}, \bm{I}^{d}_{c,i})$ using Eqs. (\ref{eq:Bd}), (\ref{eq:bd}) and (\ref{eq:LNBR}) for $i=1,2,\cdots,m$.}
    \STATE{$\bullet$ Calculate $\mathcal{L}^{m}_{\rm{NBR}}=\sum_{i=1}^{m}\mathcal{L}_{\rm{NBR}}(\bm{I}^{n}_{i}, \bm{I}^{d}_{c,i})$.}
    \STATE{}
    \STATE{\textbf{Crop-and-Paste Operation}}
    \STATE{}
    \IF{probability $>$ 50\%:}
    \STATE{$\bullet$ Calculate $\mathcal{L}_{\rm{WBCE}}(\bm{I}^{n}_{i},\bm{I}^{d}_{c,i},\bm{M}^{d}_{c,i})$ using Eqs. (\ref{eq:CaP}), (\ref{eq:LWBCE}) and (\ref{eq:lambda}) for $i=1,2,\cdots,m$.}
    \STATE{$\bullet$ Calculate $\mathcal{L}^{m}_{\rm{WBCE}}=\sum_{i=1}^{m}\mathcal{L}_{\rm{WBCE}}(\bm{I}^{n}_{i},\bm{I}^{d}_{c,i},\bm{M}^{d}_{c,i})$.}
    \STATE{$\bullet$ Calculate the overall loss of the minibatch:\\ ${\mathcal{L} }=\mathcal{L}^{m}_{\rm{NBR}}+\mathcal{L}^{m}_{\rm{WBCE}}$.}
    \STATE{}
    \ELSE
    \STATE{$\bullet$ Calculate the original binary cross-entropy loss $\mathcal{L}_{\rm{BCE}}(\bm{I}^{d}_{c,i},\bm{M}^{d}_{c,i})$ using Eq.~(\ref{eq:LBCE}).}
    \STATE{$\bullet$ Calculate $\mathcal{L}^{m}_{\rm{BCE}}=\sum_{i=1}^{m}\mathcal{L}_{\rm{BCE}}(\bm{I}^{n}_{i},\bm{I}^{d}_{c,i},\bm{M}^{d}_{c,i})$.}
    \STATE{$\bullet$ Calculate the overall loss of the minibatch:\\ ${\mathcal{L} }=\mathcal{L}^{m}_{\rm{NBR}}+\mathcal{L}^{m}_{\rm{BCE}}$.}
    \ENDIF
    \STATE{$\bullet$ Tune the network by descending the gradients of ${\mathcal{L} }$ with respect to the segmentation network parameters.}
    \ENDFOR
    \STATE{}
    \ENDFOR

\end{algorithmic}
\label{alg:training}
\end{algorithm}

\begin{figure*}[htbp]
\centering
   \includegraphics[scale=0.35]{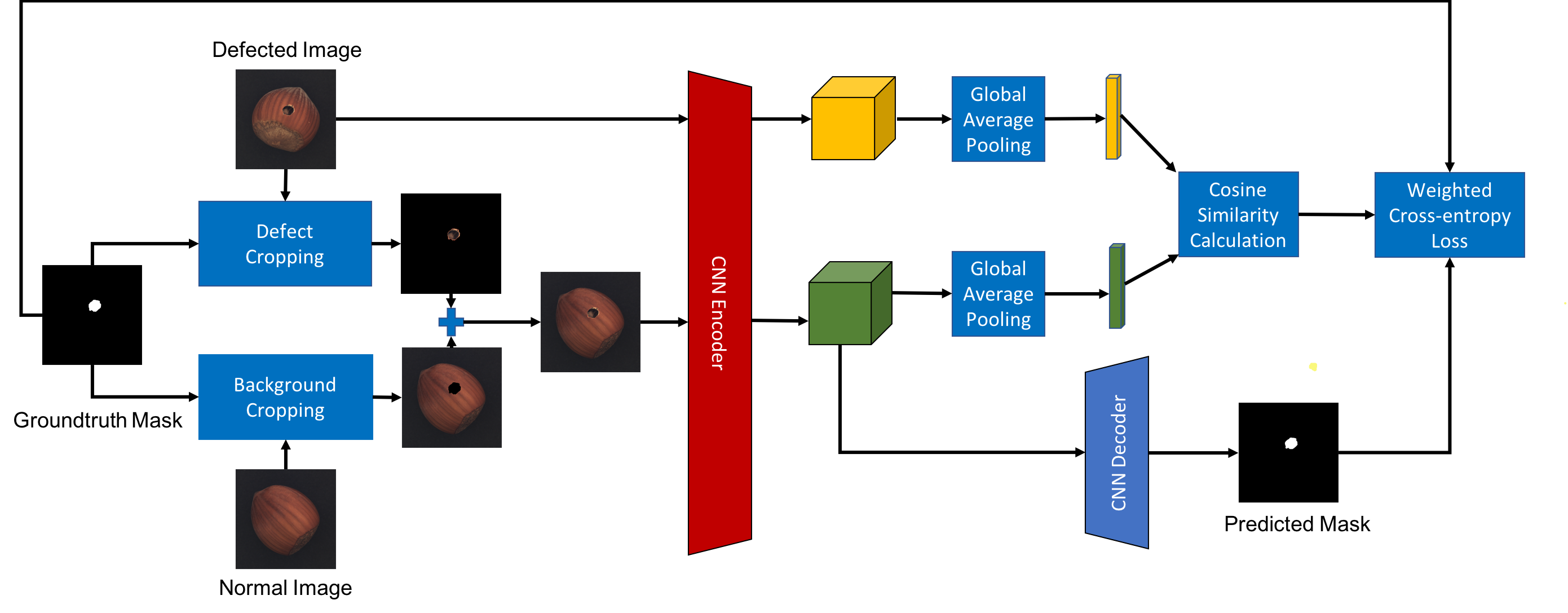}
   \caption{The illustration of Crop-and-Paste (CaP) operation.}
\label{fig:DCP}
\end{figure*}

\subsection{Crop-and-Paste (CaP)}
\label{subsec:DCP}
To further alleviate overfitting, we propose the second regularization method called $\textbf{C}$rop-\textbf{a}nd-$\textbf{P}$aste (CaP) operation inspired by Cutmix~\cite{yun2019cutmix} in image classification and object detection. As shown in Fig. \ref{fig:DCP}, the defect region in an anomalous image $\bm{I}^d$ is cropped out and pasted upon a normal image $\bm{I}^n$. Such operation is conducted using the groundtruth mask $\bm{M}^d$ by:
\begin{equation}
\bm{I}^{d}_{\rm{CaP}} = \bm{I}^d \odot \bm{M}^d + \bm{I}^n \odot (\bm{1} - \bm{M}^d).
\label{eq:CaP}
\end{equation}
The resulted augmented image $\bm{I}^{d}_{\rm{CaP}}$ is fed through the encoder and the decoder networks to generate predicted defect mask $\tilde{\bm{M}}^d_{\rm{CaP}}$ $\in [0,1]$ using sigmoid activation function on the final output layer. We propose a weighted binary cross-entropy loss by pixelwisely comparing the groundtruth mask ${\bm{M}}^d$ and $\tilde{\bm{M}}^d_{\rm{CaP}}$:
\begin{equation}
\begin{aligned}
  { \mathcal{L} }_{ \rm{WBCE} }=-\lambda\sum _{ (w,h) }^{  }\left[{ {\bm{M}}^d(w,h)log\Big(\tilde{\bm{M}}^d_{\rm{Cap}} (w,h)\Big) +} \right. \\
  \left.{\Big(1-{\bm{M}}^d(w,h)\Big)log\Big(1-\tilde{\bm{M}}^d_{\rm{CaP}} (w,h)\Big) } \right],
\end{aligned}
\label{eq:LWBCE}
\end{equation}
where $w$ and $h$ denote the width and height coordinate, respectively. The weight $\lambda$ is defined as the cosine similarity between two vectors after global average pooling $\bm{F}^d$ and $\bm{F}^{d}_{\rm{DCP}}$, i.e.,:
\begin{equation}
\lambda = \frac{|{\rm{GAP}}(\bm{F}^d)^{\rm{T}}{\rm{GAP}}(\bm{F}^{d}_{\rm{CaP}})|}{||{\rm{GAP}}(\bm{F}^d)||_2 \cdot ||{\rm{GAP}}(\bm{F}^{d}_{\rm{CaP}})||_2}.
\label{eq:lambda}
\end{equation}
$\lambda$ quantifies the contribution of ${ \mathcal{L} }_{ \rm{WBCE} }$ by the similarity between $\bm{F}^d$ and $\bm{F}^{d}_{\rm{CaP}}$ after the crop-and-paste operation. In our proposal, we simply paste the defect region to the same position as the original anomalous image rather than a random location. Such augmentation could be ``unrealistic'' depending on the picked normal image. $\lambda$ is calculated to quantify the ``realistic degree'' of the crop-and-pasted augmented image by comparing it to the original anomalous image. The augmented image is more realistic if it is highly similar to the original anomalous image. Hence, a bigger $\lambda$ reflects a relatively realistic augmented anomalous image and its loss contributes more to the tuning of the network parameters. An ablation study on the effect of $\lambda$ is conducted and analyzed in Section~\ref{subsec:ablation}.

For each minibatch iteration, we set a probability parameter which is randomly drawn uniformly from the range of [0, 1]. If the probability parameter exceeds a fixed threshold (e.g., 50\%), the CaP operation is applied. Otherwise, we feed the original feature map of anomalous image $\bm{F}^d$ through the decoder to obtain the predicted mask $\tilde{\bm{M}}^d$ and then calculate the standard binary cross-entropy loss as:
\begin{equation}
\begin{aligned}
  { \mathcal{L} }_{ \rm{BCE} }=-\sum _{ (w,h) }^{  }\left[{ {\bm{M}}^d(w,h)log\Big(\tilde{\bm{M}}^d (w,h)\Big) +} \right. \\
  \left.{\Big(1-{\bm{M}}^d(w,h)\Big)log\Big(1-\tilde{\bm{M}}^d (w,h)\Big) } \right].
\end{aligned}
\label{eq:LBCE}
\end{equation}
To update the parameters of the encoder and decoder networks, we minimize the overall loss $\mathcal{L}$ defined as the summation of $\mathcal{L}_{\rm{NBR}}$ and the segmentation loss depending on the probability parameter. Algorithm \ref{alg:training} summarizes the overall training procedure using the proposed regularization methods in pseudo codes.

\begin{remark}
Here, we provide some further analysis on NBR and CaP, respectively. Both techniques exploit a huge amount of normal training images. On one hand, NBR is formulated to train the network to generate class-wisely consistent representations of the normal regions. Without NBR, the network can only learn ``normal'' semantics from the normal regions of the very few defect training images and therefore suffers from overfitting. NBR facilitates the network to learn intra-class (normal) similarity by comparing the deep representations between the normal regions of the anomalous images and the randomly selected normal images. On the other hand, CaP is formulated to facilitate the network to learn inter-class (normal and anomalous) discriminative knowledge from different combinations of defective regions and normal images through the simple Crop-and-Paste operation. In summary, the proposed regularization techniques effectively increase the intra-class similarity for the segmentation of normal class and simultaneously enhances the inter-class discriminative capability for the normal and anomalous regions. We will study the superiority of the proposed techniques in terms of segmentation performance and model training convergence in the ablation experiments in detail in Section~\ref{subsec:ablation}.
\end{remark}



\section{Experiments}
\label{sec:experiments}

\subsection{Dataset}
\label{subsec:dataset}
\begin{figure}[!t]
  \centering
  \includegraphics[scale=0.32]{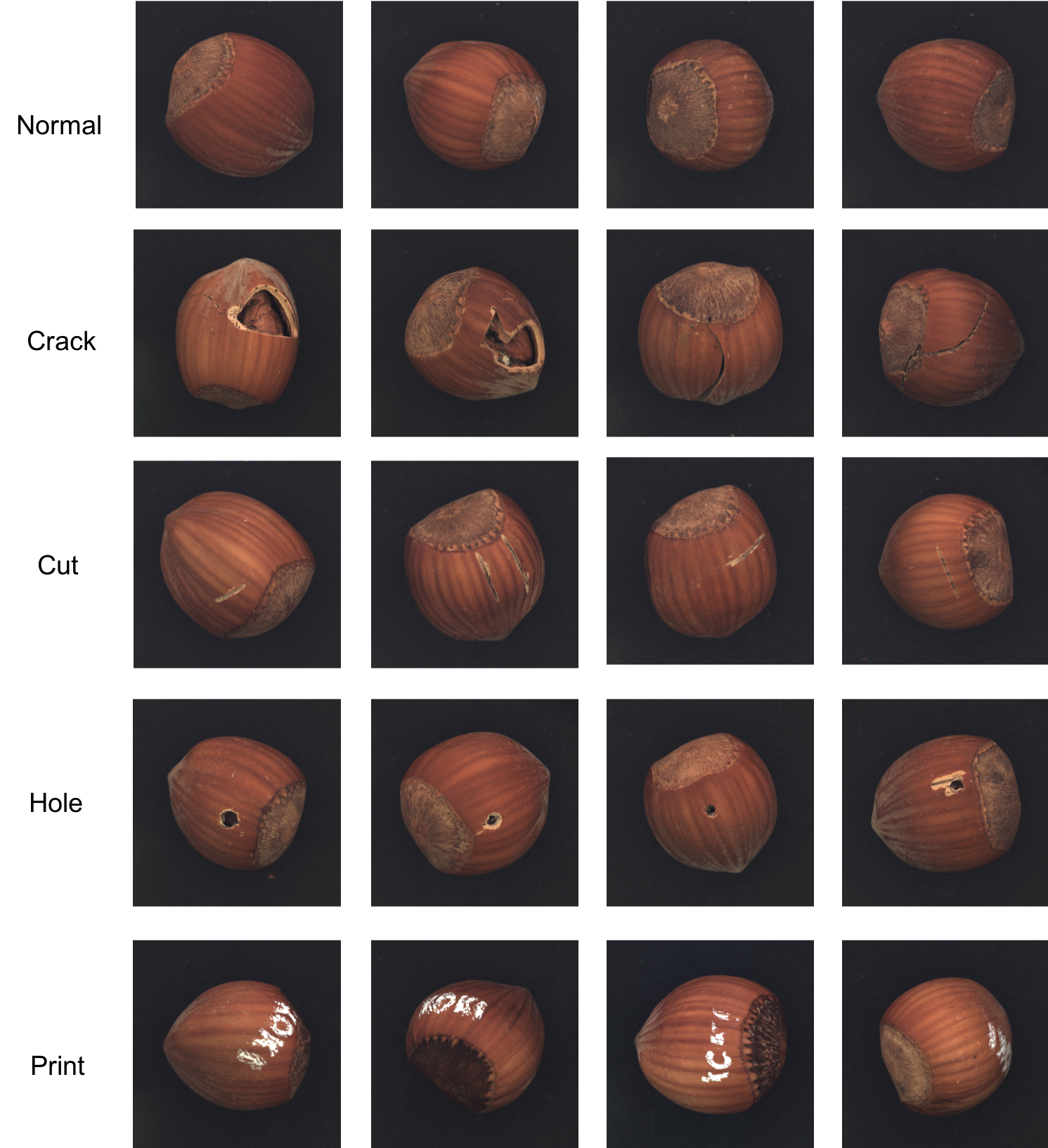}
  \caption{The examples of MVTec dataset for category ``Hazelnut''. Row 1: the normal images; Row 2 to Row 5: the anomalous images with defect ``Crack'', ``Cut'', ``Hole'' and ``Print'', respectively.}
  \label{fig:dataset}
\end{figure}

We exploit recently released MVTec Anomaly Detection (MVTec AD)~\cite{bergmann2019mvtec} dataset for experiments. MVTec AD contains 5354 high-resolution images of multiple object or texture categories. For each category, it contains defect-free images and anomalous images. Overall, MVTec AD contains over 70 different types of defects such as holes, contaminations, scratches, and etc. Fig.~\ref{fig:dataset} shows several examples of normal images and different types of anomalous images for the category ``Hazelnut''. Table~\ref{tab:dataset_info} shows the number of images for each category in the defective training subset, the normal training subset and the testing set, respectively.

\begin{table}[htbp]\scriptsize
  \centering
  \caption{The image number for each category under 1-shot and 5-shot settings}
  \scalebox{0.88}{
    \begin{tabular}{|c|c|c|c|c|c|c|}
    \toprule
    \multirow{2}[2]{*}{Category} & \multicolumn{2}{c|}{\# Train (Defect)} & \multirow{2}[2]{*}{\# Train (Normal)} & \multicolumn{2}{c|}{\# Test (Defect)} & \multirow{2}[2]{*}{\# Test (Normal)} \\
          & 1 shot & 5 shot &       & 1 shot & 5 shot &  \\
    \midrule
    Carpet & \multirow{15}[2]{*}{1} & \multirow{15}[2]{*}{5} & 280   & 88    & 84    & 28 \\
    Grid  &       &       & 264   & 56    & 52    & 21 \\
    Leather &       &       & 245   & 91    & 87    & 32 \\
    Tile  &       &       & 230   & 83    & 79    & 33 \\
    Wood  &       &       & 247   & 59    & 55    & 19 \\
    Bottle &       &       & 209   & 62    & 58    & 20 \\
    Cable &       &       & 224   & 91    & 87    & 58 \\
    Capsule &       &       & 219   & 108   & 104   & 23 \\
    Hazelnut &       &       & 391   & 69    & 65    & 40 \\
    Metal nut &       &       & 220   & 92    & 88    & 22 \\
    Pill  &       &       & 267   & 140   & 136   & 26 \\
    Screw &       &       & 320   & 118   & 114   & 41 \\
    Toothbrush &       &       & 60    & 29    & 25    & 12 \\
    Transistor &       &       & 213   & 39    & 35    & 60 \\
    Zipper &       &       & 240   & 118   & 114   & 32 \\
    \bottomrule
    \end{tabular}%
    }
  \label{tab:dataset_info}%
\end{table}%

\subsection{Experiment Settings}
We first conduct two experiments to compare our model with several benchmarking methods. The first experiment is on few-shot defect segmentation. Specifically, under K-shot (K=1 or 5) defect segmentation setting, for each type of defect, K $\left \langle \rm{image}, \rm{mask} \right \rangle$ pairs are randomly selected to construct the defect training subset while the normal training subset is constructed using the defect-free training images provided by MVTec AD. The remaining anomalous images are used for performance evaluation. The second experiment is on few-shot anomaly detection, i.e., image-level classification of normal and anomalous images under our few-shot settings. We label the normal images as category `0' while the anomalous ones as category `1'. Since the proposed method is formulated for defect segmentation, a testing image is predicted as anomalous if at least one of its pixels is predicted as defective. The anomaly score for the image is defined as the area of the defective regions. We further conduct several ablation studies to analyze the effectiveness of the proposed regularization techniques from multiple aspects.

\subsection{Benchmarking Methods}
\label{subsec:benchmarking}
We exploit the encoder-decoder architecture proposed by~\cite{buslaev2018fully} which is backboned by ResNet-34~\cite{he2016deep}. Therefore, the original model is set as baseline \textbf{B}. Our model is denoted as \textbf{B+NBR+CaP}-the baseline model with both normal background regularization and crop-and-paste operation. To build a strong baseline \textbf{B}, the similar crop-and-paste data augmentation is applied at the probability of 50\% for randomly sampled minibatches and the standard binary cross-entropy loss (without weighting) is used for optimization. We also conduct the experiments on TernausNet-11~\cite{iglovikov2018ternausnet}, TernausNet-16~\cite{shvets2018automatic}, the original U-Net~\cite{ronneberger2015u}, Attention U-Net~\cite{oktay2018attention} and U-Net++~\cite{zhou2019unet++} as benchmarking methods. For all the benchmarking methods, we also exploit crop-and-paste data augmentation but they are trained using the standard binary cross-entropy loss.


\begin{table*}[htbp]\scriptsize
  \centering
  \caption{Mean IOU under 1-shot and 5-shot settings. The best values are in bold.}
  \begin{threeparttable}
    \begin{tabular}{|cc|c|c|c|c|c|c|c|}
    \toprule
    \multicolumn{1}{|c|}{\multirow{2}[2]{*}{}} & \multirow{2}[2]{*}{Category} & \multirow{2}[2]{*}{TernausNet-11~\cite{iglovikov2018ternausnet}} & \multirow{2}[2]{*}{TernausNet-16~\cite{shvets2018automatic}} & \multirow{2}[2]{*}{U-Net~\cite{ronneberger2015u}} & \multirow{2}[2]{*}{Attention U-Net~\cite{oktay2018attention}} & \multirow{2}[2]{*}{U-Net++~\cite{zhou2019unet++}} & \multirow{2}[2]{*}{B (AlbuNet)~\cite{buslaev2018fully}} & \multirow{2}[2]{*}{B+NBR+CaP (ours)} \\
    \multicolumn{1}{|c|}{} &       &       &       &       &       &       &       &  \\
    \midrule
    \multicolumn{1}{|c|}{\multirow{15}[4]{*}{1-shot}} & Carpet & 0.4445(0.0409) & 0.4699(0.0192) & 0.4995(0.0035) & 0.4479(0.0476) & 0.4931(0.0163) & 0.3461(0.0757) & \textbf{0.5854}(0.0069) \\
    \multicolumn{1}{|c|}{} & Grid  & 0.2787(0.0046) & 0.2320(0.0301) & 0.2787(0.0208) & 0.2211(0.0026) & 0.2904(0.0216) & 0.2090(0.0408) & \textbf{0.3364}(0.0075) \\
    \multicolumn{1}{|c|}{} & Leather & 0.3987(0.0765) & 0.1453(0.0612) & 0.3154(0.0103) & 0.1880(0.0432) & 0.2673(0.0156) & 0.4589(0.0057) & \textbf{0.5578}(0.0040) \\
    \multicolumn{1}{|c|}{} & Tile  & 0.6873(0.0097) & 0.6772(0.0089) & 0.7522(0.0100) & 0.6761(0.0175) & 0.7519(0.0185) & 0.7557(0.0217) & \textbf{0.8377}(0.0138) \\
    \multicolumn{1}{|c|}{} & Wood  & 0.4803(0.0366) & 0.4834(0.0832) & 0.5008(0.0267) & 0.4274(0.0393) & 0.4922(0.0136) & 0.4600(0.0304) & \textbf{0.6004}(0.0079) \\
\cmidrule{2-9}    \multicolumn{1}{|c|}{} & Bottle & 0.2988(0.0033) & 0.2867(0.0144) & 0.2655(0.0251) & 0.2908(0.0253) & 0.2815(0.0196) & 0.3291(0.0327) & \textbf{0.6763}(0.0152) \\
    \multicolumn{1}{|c|}{} & Cable & 0.2412(0.0164) & 0.2338(0.0164) & 0.2434(0.0070) & 0.2731(0.0145) & 0.2375(0.0202) & 0.2482(0.0193) & \textbf{0.3197}(0.0048) \\
    \multicolumn{1}{|c|}{} & Capsule & 0.1316(0.0024) & 0.1316(0.0024) & 0.1101(0.0260) & 0.1107(0.0269) & 0.2143(0.0088) & 0.1346(0.0030) & \textbf{0.2631}(0.0046) \\
    \multicolumn{1}{|c|}{} & Hazelnut & 0.3556(0.0176) & 0.3102(0.0541) & 0.3266(0.0438) & 0.2874(0.0679) & 0.4560(0.0769) & 0.4212(0.0339) & \textbf{0.6671}(0.0089) \\
    \multicolumn{1}{|c|}{} & Metal nut & 0.2465(0.0464) & 0.2958(0.0089) & 0.1876(0.0677) & 0.1551(0.0576) & 0.3920(0.0709) & 0.4647(0.0504) & \textbf{0.6234}(0.0105) \\
    \multicolumn{1}{|c|}{} & Pill  & 0.1475(0.0132) & 0.1254(0.0434) & 0.1506(0.0191) & 0.1478(0.0225) & 0.1922(0.0090) & 0.1970(0.0140) & \textbf{0.4245}(0.0402) \\
    \multicolumn{1}{|c|}{} & Screw & 0.0918(0.0062) & 0.0894(0.0237) & 0.0937(0.0853) & 0.0857(0.0076) & 0.0708(0.0058) & 0.0482(0.0629) & \textbf{0.1801}(0.0398) \\
    \multicolumn{1}{|c|}{} & Toothbrush\tnote{*} & -    &  -    & -    & -    & -    & -    & - \\
    \multicolumn{1}{|c|}{} & Transistor & 0.1113(0.0018) & 0.1425(0.0064) & 0.1314(0.0327) & 0.1516(0.0550) & 0.1214(0.0057) & 0.1193(0.0360) & \textbf{0.2061}(0.0081) \\
    \multicolumn{1}{|c|}{} & Zipper & 0.4771(0.0561) & 0.4597(0.0175) & 0.5339(0.0397) & 0.4905(0.0119) & 0.5804(0.0325) & 0.4651(0.0325) & \textbf{0.6096}(0.0056) \\
    \midrule
    \multicolumn{2}{|c|}{Mean} & 0.3137 & 0.2917 & 0.3136 & 0.2824 & 0.3458 & 0.3327 & \textbf{0.4919} \\
    \midrule
    \multicolumn{1}{|c|}{\multirow{15}[4]{*}{5-shot}} & Carpet & 0.5934(0.0115) & 0.5902(0.0084) & 0.6692(0.0054) & 0.5292(0.1215) & 0.6331(0.0034) & 0.5938(0.0069) & \textbf{0.7112}(0.0180) \\
    \multicolumn{1}{|c|}{} & Grid  & 0.5023(0.0083) & 0.4189(0.0134) & 0.5412(0.0021) & 0.4677(0.0264) & 0.5543(0.0291) & 0.4514(0.0343) & \textbf{0.5577}(0.0040) \\
    \multicolumn{1}{|c|}{} & Leather & 0.4564(0.0355) & 0.4084(0.0339) & 0.5788(0.0455) & 0.4150(0.0018) & 0.5489(0.0316) & 0.6062(0.0169) & \textbf{0.6887}(0.0025) \\
    \multicolumn{1}{|c|}{} & Tile  & 0.8203(0.0047) & 0.7611(0.1160) & 0.8498(0.0059) & 0.8104(0.0109) & 0.8439(0.0157) & 0.8505(0.0021) & \textbf{0.8713}(0.0033) \\
    \multicolumn{1}{|c|}{} & Wood  & 0.6265(0.0003) & 0.6491(0.0303) & 0.6288(0.0114) & 0.6225(0.0120) & 0.7363(0.0029) & 0.6856(0.0048) & \textbf{0.7367}(0.0231) \\
\cmidrule{2-9}    \multicolumn{1}{|c|}{} & Bottle & 0.4799(0.0528) & 0.5737(0.0112) & 0.4991(0.0015) & 0.4551(0.0014) & 0.6117(0.0109) & 0.5456(0.0830) & \textbf{0.7615}(0.0035) \\
    \multicolumn{1}{|c|}{} & Cable & 0.4848(0.0001) & 0.4334(0.0334) & 0.4470(0.0183) & 0.4407(0.0421) & 0.4750(0.0021) & 0.4835(0.0110) & \textbf{0.5891}(0.0190) \\
    \multicolumn{1}{|c|}{} & Capsule & 0.3067(0.0018) & 0.2913(0.0035) & 0.3187(0.0112) & 0.3285(0.0340) & 0.3224(0.0117) & 0.3408(0.0028) & \textbf{0.4814}(0.0242) \\
    \multicolumn{1}{|c|}{} & Hazelnut & 0.6894(0.0173) & 0.6955(0.0016) & 0.7037(0.0218) & 0.6621(0.0262) & 0.7618(0.0260) & 0.7542(0.0238) & \textbf{0.7876}(0.0142) \\
    \multicolumn{1}{|c|}{} & Metal nut & 0.6451(0.0640) & 0.6763(0.0004) & 0.5385(0.1620) & 0.6649(0.0131) & 0.7244(0.0006) & 0.6540(0.0805) & \textbf{0.7742}(0.0158) \\
    \multicolumn{1}{|c|}{} & Pill  & 0.5035(0.0668) & 0.4749(0.0165) & 0.4610(0.0351) & 0.4222(0.0436) & 0.5629(0.0401) & 0.3623(0.0529) & \textbf{0.6976}(0.0163) \\
    \multicolumn{1}{|c|}{} & Screw & 0.2512(0.0014) & 0.2533(0.0544) & 0.2351(0.0109) & 0.2385(0.0142) & 0.2628(0.0316) & 0.2204(0.0385) & \textbf{0.3696}(0.0364) \\
    \multicolumn{1}{|c|}{} & Toothbrush & 0.0113(0.0117) & 0.0627(0.0403) & 0.0197(0.0006) & 0.0234(0.0128) & 0.0752(0.0212) & 0.1382(0.0380) & \textbf{0.2982}(0.0328) \\
    \multicolumn{1}{|c|}{} & Transistor & 0.4317(0.0350) & 0.4345(0.0906) & 0.3754(0.0749) & 0.4145(0.0345) & 0.4253(0.0537) & 0.3164(0.0689) & \textbf{0.5766}(0.0144) \\
    \multicolumn{1}{|c|}{} & Zipper & 0.6214(0.0105) & 0.5928(0.0057) & 0.6171(0.0052) & 0.6002(0.0074) & 0.6560(0.0247) & 0.6142(0.0064) & \textbf{0.6786}(0.0065) \\
    \midrule
    \multicolumn{2}{|c|}{Mean} & 0.4949 & 0.4877 & 0.4989 & 0.4730 & 0.5463 & 0.5078 & \textbf{0.6296} \\
    \bottomrule
    \end{tabular}%

  \begin{tablenotes}
    \item[*] For category toothbrush, all the compared methods cannot converge during training under 1-shot setting.
  \end{tablenotes}

  \end{threeparttable}
  \label{tab:5_time_iou}%
\end{table*}%

\begin{table*}[htbp]\scriptsize
  \centering
  \caption{Mean DC under 1-shot and 5-shot settings. The best values are in bold.}
  \begin{threeparttable}
    \begin{tabular}{|cc|c|c|c|c|c|c|c|}
    \toprule
    \multicolumn{1}{|c|}{\multirow{2}[2]{*}{}} & \multirow{2}[2]{*}{Category} & \multirow{2}[2]{*}{TernausNet-11~\cite{iglovikov2018ternausnet}} & \multirow{2}[2]{*}{TernausNet-16~\cite{shvets2018automatic}} & \multirow{2}[2]{*}{U-Net~\cite{ronneberger2015u}} & \multirow{2}[2]{*}{Attention U-Net~\cite{oktay2018attention}} & \multirow{2}[2]{*}{U-Net++~\cite{zhou2019unet++}} & \multirow{2}[2]{*}{B (AlbuNet)~\cite{buslaev2018fully}} & \multirow{2}[2]{*}{B+NBR+CaP (ours)} \\
    \multicolumn{1}{|c|}{} &       &       &       &       &       &       &       &  \\
    \midrule
    \multicolumn{1}{|c|}{\multirow{15}[4]{*}{1-shot}} & Carpet & 0.5721(0.0433 & 0.5857(0.0603) & 0.6317(0.0039) & 0.5728(0.0528) & 0.6267(0.0210) & 0.4690(0.0915) & \textbf{0.7012}(0.0052) \\
    \multicolumn{1}{|c|}{} & Grid  & 0.4080(0.0016) & 0.3278(0.0381) & 0.4055(0.0309) & 0.3393(0.0022) & 0.4154(0.0247) & 0.3134(0.0549) & \textbf{0.4688}(0.0112) \\
    \multicolumn{1}{|c|}{} & Leather & 0.5294(0.0858) & 0.2587(0.0726) & 0.4365(0.0100) & 0.2716(0.0604) & 0.3705(0.0206) & 0.5839(0.0106) & \textbf{0.6853}(0.0046) \\
    \multicolumn{1}{|c|}{} & Tile  & 0.7789(0.0081) & 0.7691(0.0059) & 0.8368(0.0101) & 0.7711(0.0163) & 0.8375(0.0134) & 0.8476(0.0170) & \textbf{0.9050}(0.0107) \\
    \multicolumn{1}{|c|}{} & Wood  & 0.6217(0.0395) & 0.6796(0.0837) & 0.6403(0.0229) & 0.5671(0.0440) & 0.6283(0.0123) & 0.5859(0.0289) & \textbf{0.7253}(0.0083) \\
\cmidrule{2-9}    \multicolumn{1}{|c|}{} & Bottle & 0.3990(0.0036) & 0.3973(0.0117) & 0.3577(0.0385) & 0.3898(0.0301) & 0.3696(0.0270) & 0.4421(0.0398) & \textbf{0.7880}(0.0132) \\
    \multicolumn{1}{|c|}{} & Cable & 0.2900(0.0316) & 0.2532(0.0245) & 0.2795(0.0094) & 0.3283(0.0240) & 0.2705(0.0275) & 0.2823(0.0273) & \textbf{0.3850}(0.0078) \\
    \multicolumn{1}{|c|}{} & Capsule & 0.1961(0.0024) & 0.1944(0.0024) & 0.1660(0.0384) & 0.1654(0.0337) & 0.2986(0.0191) & 0.1983(0.0015) & \textbf{0.3531}(0.0072) \\
    \multicolumn{1}{|c|}{} & Hazelnut & 0.4767(0.0223) & 0.3749(0.0644) & 0.4377(0.0508) & 0.3871(0.0793) & 0.5701(0.0844) & 0.5290(0.0362) & \textbf{0.7752}(0.0100) \\
    \multicolumn{1}{|c|}{} & Metal nut & 0.3142(0.0535) & 0.3460(0.0128) & 0.2478(0.0793) & 0.2034(0.0481) & 0.4933(0.0655) & 0.5346(0.0579) & \textbf{0.7195}(0.0131) \\
    \multicolumn{1}{|c|}{} & Pill  & 0.2032(0.0179) & 0.2124(0.0544) & 0.2027(0.0233) & 0.2027(0.0282) & 0.2501(0.0122) & 0.2547(0.0106) & \textbf{0.5254}(0.0454) \\
    \multicolumn{1}{|c|}{} & Screw & 0.1297(0.0080) & 0.1599(0.0363) & 0.1815(0.0387) & 0.1200(0.0049) & 0.0980(0.0035) & 0.0816(0.1061) & \textbf{0.2248}(0.0541) \\
    \multicolumn{1}{|c|}{} & Toothbrush\tnote{*} & -    & -    & -    & -    & -    & -    & - \\
    \multicolumn{1}{|c|}{} & Transistor & 0.1715(0.0089) & 0.1905(0.0120) & 0.1627(0.0754) & 0.2206(0.0735) & 0.1768(0.0112) & 0.1754(0.0421) & \textbf{0.2840}(0.0063) \\
    \multicolumn{1}{|c|}{} & Zipper & 0.6251(0.0583) & 0.6256(0.0209) & 0.6824(0.0383) & 0.6418(0.0164) & 0.7214(0.0307) & 0.6057(0.0349) & \textbf{0.7416}(0.0054) \\
    \midrule
    \multicolumn{2}{|c|}{Mean} & 0.4083 & 0.3839 & 0.4049 & 0.3701 & 0.4377 & 0.4217 & \textbf{0.5916} \\
    \midrule
    \multicolumn{1}{|c|}{\multirow{15}[4]{*}{5-shot}} & Carpet & 0.7225(0.0164) & 0.7195(0.0069) & 0.7864(0.0057) & 0.6621(0.1039) & 0.7534(0.0001) & 0.7281(0.0068) & \textbf{0.8219}(0.0146) \\
    \multicolumn{1}{|c|}{} & Grid  & 0.6490(0.0074) & 0.5681(0.0086) & 0.6922(0.0021) & 0.6187(0.0255) & 0.6968(0.0262) & 0.6020(0.0320) & \textbf{0.6852}(0.0046) \\
    \multicolumn{1}{|c|}{} & Leather & 0.6012(0.0315) & 0.5591(0.0357) & 0.7121(0.0392) & 0.5603(0.0026) & 0.6829(0.0347) & 0.7304(0.0205) & \textbf{0.8033}(0.0018) \\
    \multicolumn{1}{|c|}{} & Tile  & 0.8948(0.0023) & 0.8422(0.0977) & 0.9145(0.0053) & 0.8874(0.0079) & 0.9102(0.0107) & 0.9153(0.0003) & \textbf{0.9282}(0.0021) \\
    \multicolumn{1}{|c|}{} & Wood  & 0.7523(0.0027) & 0.7738(0.0270) & 0.7480(0.0078) & 0.7522(0.0082) & 0.8403(0.0030) & 0.8049(0.0057) & \textbf{0.8410}(0.0182) \\
\cmidrule{2-9}    \multicolumn{1}{|c|}{} & Bottle & 0.6088(0.0511) & 0.6905(0.0100) & 0.6250(0.0072) & 0.5860(0.0011) & 0.7291(0.0064) & 0.6586(0.0784) & \textbf{0.8490}(0.0033) \\
    \multicolumn{1}{|c|}{} & Cable & 0.5877(0.0036) & 0.5261(0.0286) & 0.5416(0.0202) & 0.5403(0.0361) & 0.5699(0.0037) & 0.5699(0.0140) & \textbf{0.6768}(0.0156) \\
    \multicolumn{1}{|c|}{} & Capsule & 0.4199(0.0047) & 0.4037(0.0019) & 0.4306(0.0101) & 0.4375(0.0419) & 0.4311(0.0122) & 0.4565(0.0033) & \textbf{0.6035}(0.0250) \\
    \multicolumn{1}{|c|}{} & Hazelnut & 0.7999(0.0124) & 0.7986(0.0018) & 0.8091(0.0139) & 0.7801(0.0235) & 0.8504(0.0252) & 0.8489(0.0212) & \textbf{0.8762}(0.0097) \\
    \multicolumn{1}{|c|}{} & Metal nut & 0.7381(0.0706) & 0.7689(0.0001) & 0.6365(0.1570) & 0.7570(0.0062) & 0.8108(0.0027) & 0.7516(0.0754) & \textbf{0.8574}(0.0150) \\
    \multicolumn{1}{|c|}{} & Pill  & 0.6240(0.0673) & 0.5865(0.0161) & 0.5795(0.0361) & 0.5366(0.0448) & 0.6745(0.0453) & 0.4620(0.0547) & \textbf{0.8022}(0.0180) \\
    \multicolumn{1}{|c|}{} & Screw & 0.3235(0.0095) & 0.3303(0.0670) & 0.3006(0.0129) & 0.3142(0.0158) & 0.3257(0.0307) & 0.2795(0.0467) & \textbf{0.4458}(0.0414) \\
    \multicolumn{1}{|c|}{} & Toothbrush & 0.0165(0.0225) & 0.0848(0.0556) & 0.0309(0.0006) & 0.0372(0.0172) & 0.1085(0.0293) & 0.2014(0.0474) & \textbf{0.4060}(0.0349) \\
    \multicolumn{1}{|c|}{} & Transistor & 0.5421(0.0513) & 0.5456(0.0981) & 0.4937(0.0935) & 0.5354(0.0407) & 0.5587(0.0593) & 0.4072(0.0783) & \textbf{0.6981}(0.0180) \\
    \multicolumn{1}{|c|}{} & Zipper & 0.7542(0.0096) & 0.7283(0.0061) & 0.7483(0.0045) & 0.7358(0.0056) & 0.7803(0.0195) & 0.7481(0.0009) & \textbf{0.7977}(0.0058) \\
    \midrule
    \multicolumn{2}{|c|}{Mean} & 0.6023 & 0.5950 & 0.6033 & 0.5827 & 0.6482 & 0.6110 & \textbf{0.7320} \\
    \bottomrule
    \end{tabular}%

  \begin{tablenotes}
    \item[*] For category toothbrush, all the compared methods cannot converge during training under 1-shot setting.
  \end{tablenotes}

  \end{threeparttable}
  \label{tab:5_time_dc}%
\end{table*}%

\begin{figure*}[htbp]
  \centering
  \includegraphics[scale=0.3]{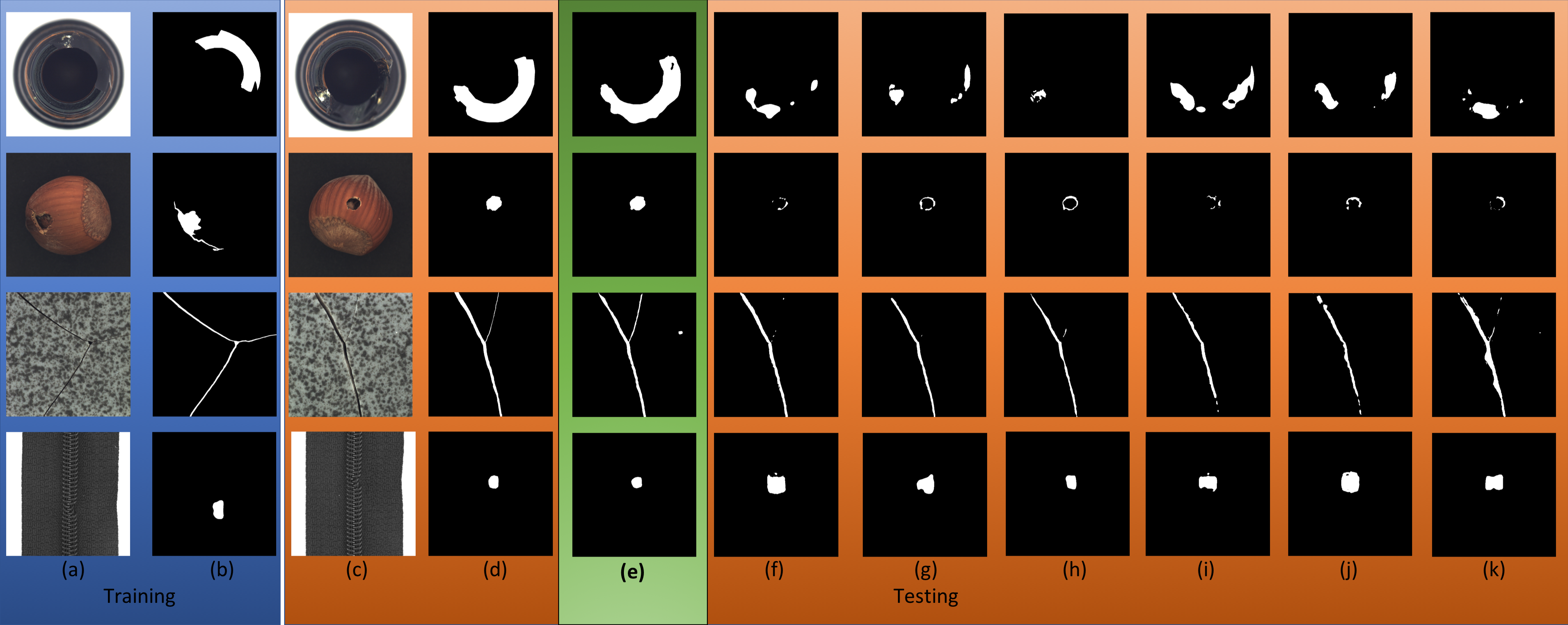}\\
  \caption{The qualitative segmentation results of category ``Bottle'', ``Hazelnut'', ``Tile'' and ``Zipper'' under 1-shot defect segmentation setting. (a) and (b): the training image and its groundtruth; (c) and (d): the testing image and its groundtruth; (e) to (i): the predicted masks of (e) our method (B+NBR+CaP); (f) B~\cite{buslaev2018fully}; (g) TernausNet-11~\cite{iglovikov2018ternausnet}; (h) TernausNet-16~\cite{shvets2018automatic}; (i) U-Net~\cite{ronneberger2015u}; (j) Attention U-Net~\cite{oktay2018attention}; (k) U-Net++~\cite{zhou2019unet++}, respectively.}
  \label{fig:qualitative}
\end{figure*}

\subsection{Performance Metrics}
\label{subsec:performance_metrics}
For the experiments of few-shot defect segmentation and the ablation studies, the performance criteria Intersection-Over-Union (IOU) and Dice Coefficient (DC) are adopted to evaluate the segmentation performance. IOU and DC are adopted to measure the overlapping regions between the predicted binary mask and the groundtruth:

\begin{equation}
\begin{array}{l}
\rm{IOU = \frac{TP}{TP + FN + FP}},\\
\\
\rm{DC = \frac{2TP}{2TP + FN + FP}}.

\end{array}
\end{equation}\label{BoundingBoxCoordinates}

\noindent where TP, FP, and FN denotes the number of pixels in the predicted mask which are true positive, false positive and false negative compared with the groundtruth mask, respectively.  The mean IOU and DC are recorded over the testing images.

For the experiments of few-shot anomaly detection, we plot the Receiver Operating Characteristic (ROC) curves for the compared methods under 1-shot and 5-shot setting, respectively. The average classification accuracy and the area under the ROC curves are calculated as the performance metrics for evaluation.


\subsection{Implementation Details}
\label{subsec:implementation}
The experiments are implemented using Pytorch~\cite{adam2017automatic} framework and Python3 ran on a NVIDIA TITAN X GPU. The training and testing images (along with their annotation masks) are resized to 512$\times$512 in resolution. All the compared methods have two branches of input images, i.e., the input branches for the normal training subset and the defect training subset, respectively. For both branches, the batch size is set as 2 for 1-shot and 4 for 5-shot defect segmentation, respectively. Adam optimizer with $\beta_1=0.9$, $\beta_2=0.999$ are exploited and the learning rate is set as 0.0001. The number of training iterations is set as 1350. For our proposed method, the probability of adopting the crop-and-paste is set as $50\%$ and for the rest of the tuning iterations, we minimize the standard binary cross-entropy loss with respect to the original anomalous training images without CaP operation. The same probability parameter for crop-and-paste operation is applied for the benchmarking methods. It is noted that all the proposed regularization methods are applied only in the training process. For all the benchmarking methods, we use the same batch size, optimizer, learning rate and the number of the iterations as our method since we can observe the convergence of training loss under these settings.



\subsection{Comparison on Few-Shot Defect Segmentation}
\label{subsec:comparison_few_shot_seg}



In this subsection, we conduct the experiments on the comparison among our models and the related benchmarking methods in few-shot defect segmentation. For each model, we repeat the experiment for 5 times with different random seeds. Table~\ref{tab:5_time_iou} and \ref{tab:5_time_dc} show the mean IOU and DC and the standard deviation for each category under 1-shot and 5-shot setting, respectively. It is observed that the proposed method significantly outperforms all the benchmarking methods in terms of mean IOU and mean DC under both few-shot settings. The performance gain shrinks from 1-shot to 5-shot because the overfitting is alleviated by increasing the number of anomalous images. Note that even when more annotated training data is provided, our method could still improve the segmentation performance. It is worthwhile to note that the original U-Net beats TernausNet-11, TernausNet-16 and \textbf{B} under 1-shot setting but is comparable or slightly underperforms them under 5-shot setting. This indicates the lower capacity model (i.e., with lower number of tunable parameters) shows better generalization performance than the higher ones in the case of extremely limited annotated training data. From the experimental results, the proposed methods effectively regularizes a high capacity baseline model \textbf{B} and significantly boost the segmentation performance. Fig.~\ref{fig:qualitative} shows the qualitative segmentation results of compared methods under 1-shot setting. We choose two representative categories for ``Object'' and ``Texture'', respectively. It is observed that, providing only a single annotated training image, our model could produce better generalization segmentation performance than all the benchmarking methods, especially on the fine details such as object edges and thin cracks. To summarize, both quantitatively and qualitatively, the proposed method significantly outperforms all the benchmarking methods under both few-shot settings.

\subsection{Comparison on Few-Shot Anomaly Detection}
\label{subsec:comparison_few_shot_ad}
\begin{figure}[htpb]
   \subfloat[][]{
      \includegraphics[scale=0.55]{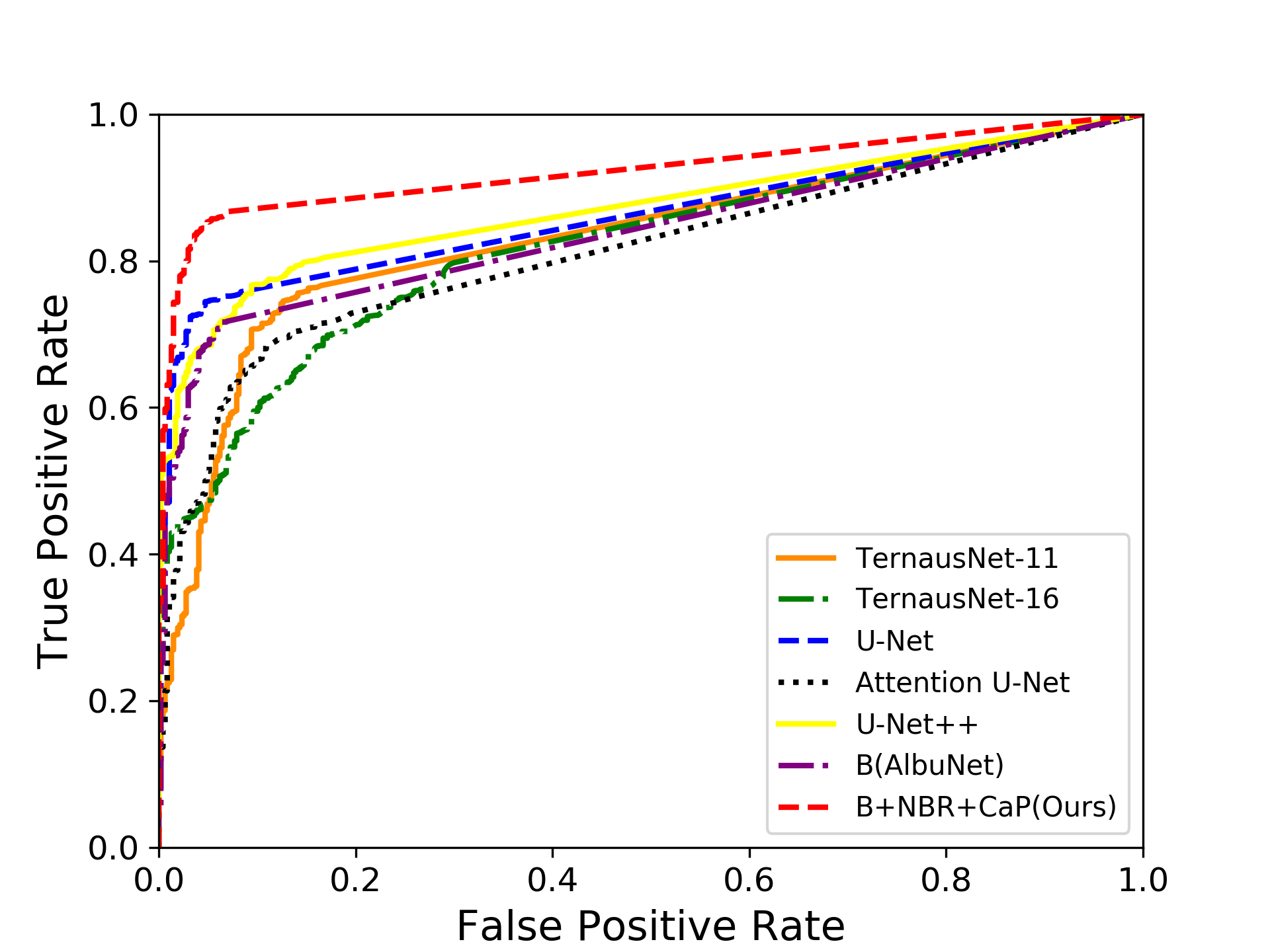}
      \centering
      \label{1_shot_roc}
   }\\
   \subfloat[][]{
      \includegraphics[scale=0.55]{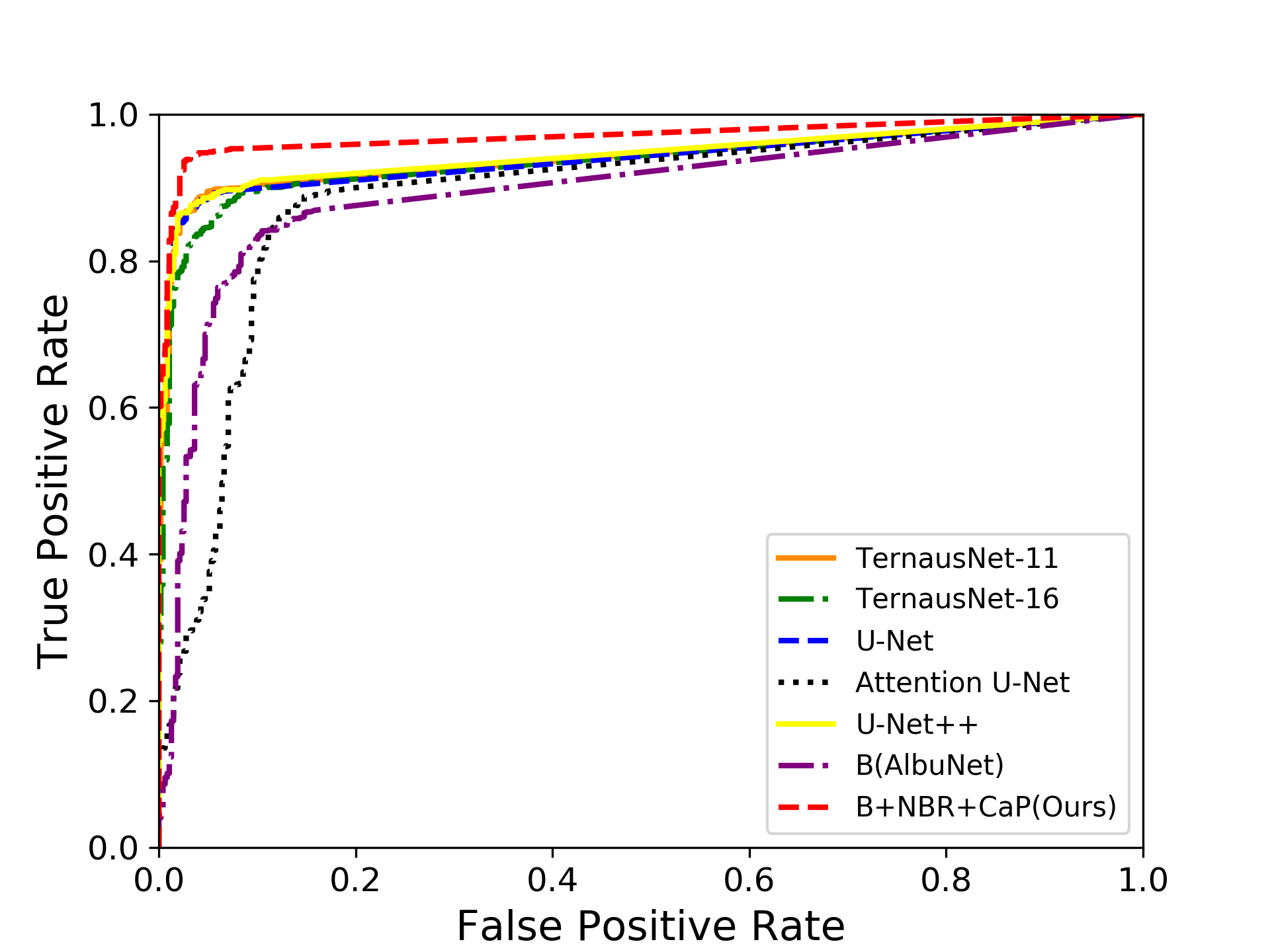}
      \centering
      \label{5_shot_roc}
   }
   \label{fig:roc_curves}%
   \caption{The ROC curves of the compared methods under (a) 1-shot and (b) 5-shot settings.}
\end{figure}

In this subsection, we conduct the experiments on few-shot anomaly detection. Since all the compared methods are originally formulated for few-shot defect segmentation, we determine a testing image to be anomalous if at least one of its pixels is predicted as defective. The anomaly score for the image is defined as the area of the defective regions. The ROC curves are plotted by thresholding the anomaly scores. Fig~\ref{1_shot_roc} and \ref{5_shot_roc} illustrates the ROC curves for the compared methods under 1-shot and 5-shot setting, respectively. Table~\ref{tab:acc_and_auc} records the average classification accuracy and the areas under the ROC curves, respectively. It is noted that the ROC curves and the performance metrics are averaged across all the categories in MVTec dataset. From the ROC curves, it is observed that our proposed model outperforms all the benchmarking methods under both settings. From the table, our model produces higher classification accuracy than all the benchmarking methods. Specifically, under 1-shot setting, the proposed model produces 88.56\% and 92.19\% in classification accuracy and AUC value, respectively. Under 5-shot setting, we reach 94.12\% and 97.03\%. We achieve such high classification accuracy at very low annotation cost by means of leveraging sufficient normal training data which is annotation-free.

\begin{table}[htbp]
  \centering
  \caption{Classification accuracy (ACC) and area under ROC curves (AUC) under 1-shot and 5-shot settings. The best values are in bold.}
    \begin{tabular}{|c|cc|cc|}
    \toprule
    \multirow{2}[2]{*}{Method} & \multicolumn{2}{c|}{1 shot} & \multicolumn{2}{c|}{5 shot} \\
          & ACC   & AUC   & ACC   & AUC \\
    \midrule
    \midrule
    TernausNet-11~\cite{iglovikov2018ternausnet} & 78.57\% & 83.13\% & 90.51\% & 94.11\% \\
    TernausNet-16~\cite{shvets2018automatic} & 76.99\% & 81.83\% & 89.04\% & 93.58\% \\
    U-Net~\cite{ronneberger2015u} & 80.26\% & 86.25\% & 90.88\% & 93.98\% \\
    Attention U-Net~\cite{oktay2018attention} & 74.94\% & 81.22\% & 86.69\% & 88.84\% \\
    U-Net++~\cite{zhou2019unet++} & 81.23\% & 87.10\% & 90.01\% & 94.47\% \\
    B(AlbuNet)~\cite{buslaev2018fully} & 77.78\% & 83.97\% & 85.96\% & 89.47\% \\
    \midrule
    B+NBR+CaP (ours) & \textbf{88.56\%} & \textbf{92.19\%} & \textbf{94.12\%} & \textbf{97.03\%} \\
    \bottomrule
    \end{tabular}%
  \label{tab:acc_and_auc}%
\end{table}%


\subsection{Ablation Study and Discussion}
\label{subsec:ablation}
In this subsection, some ablation studies and discussions are conducted on the proposed methods. We mainly consider three ablation models: \textbf{B}-the baseline model; \textbf{B+NBR}-the baseline model with normal background regularization; \textbf{B+NBR+CaP}-the baseline model with both normal background regularization and crop-and-paste operation.

\begin{figure}[!t]
  \centering
  \includegraphics[scale=0.3]{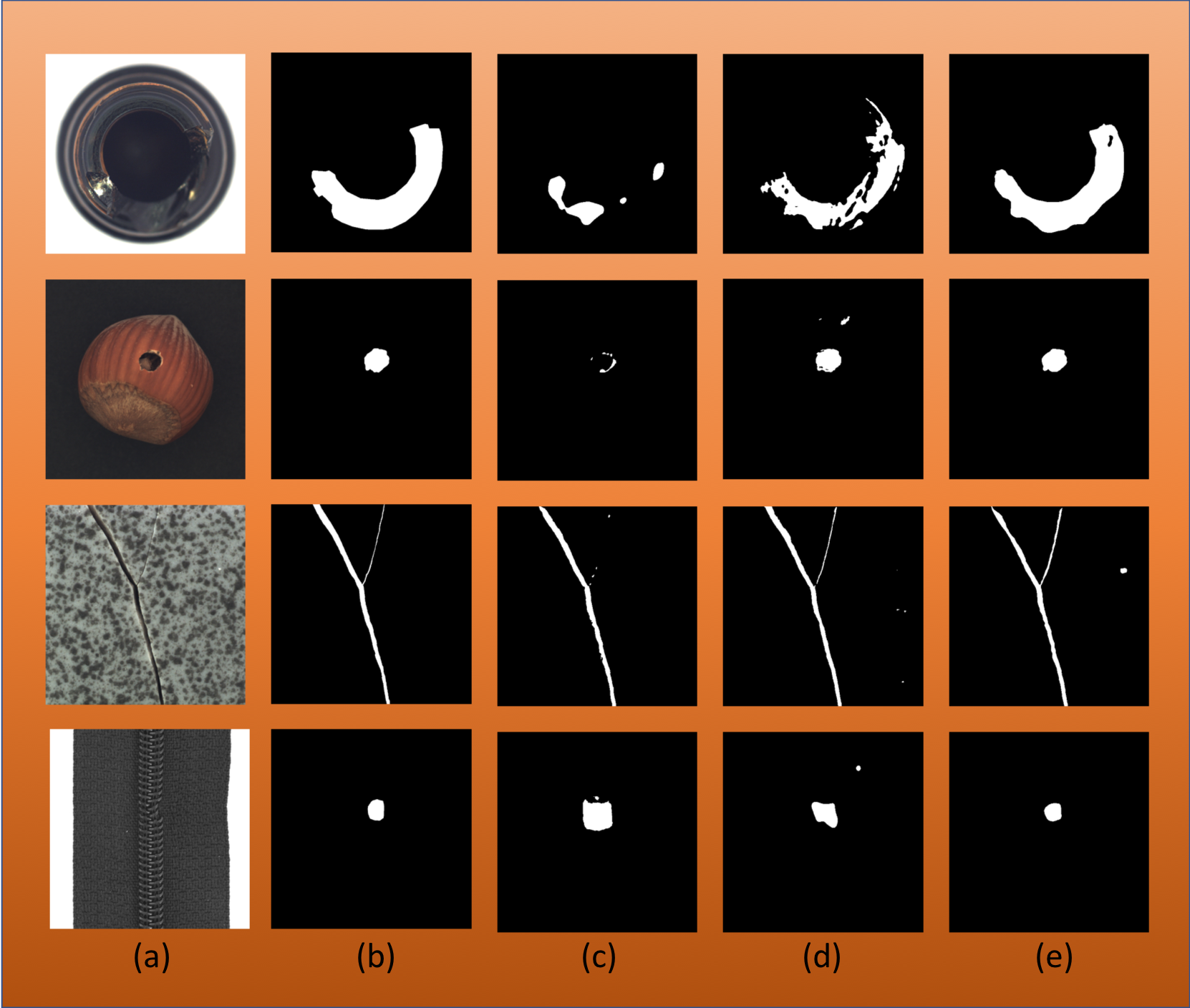}\\
  \caption{The qualitative segmentation results of category ``Bottle'', ``Hazelnut'', ``Tile'' and ``Zipper'' under 1-shot defect segmentation setting. (a) and (b): the testing image and its groundtruth; (c) to (e): the predicted masks of (c) B~\cite{buslaev2018fully}; (d) B+NBR; (e) B+NBR+CaP, respectively.}
  \label{fig:qualitative_ablation}
\end{figure}

Table~\ref{tab:ablation_IOU_DICE} shows the mean IOUs and DCs over all the categories for three ablation models under 1-shot and 5-shot settings, respectively. From the table, it is clearly that both \textbf{B+NBR} and \textbf{B+NBR+CaP} significantly outperform the original baseline model \textbf{B} by big margin in both evaluation metrics. Secondly, \textbf{B+NBR+CaP} further boosts the performance of \textbf{B+NBR} by around 7\% and 4\% under 1-shot and 5-shot settings, respectively. Fig.~\ref{fig:qualitative_ablation} shows the qualitative segmentation examples under 1-shot setting. It is observed that \textbf{B+NBR} can better reduce the segmentation errors on the normal background regions by conducting normal background regularization and \textbf{B+NBR+CaP} further improves the segmentation performance on the defect regions. These observations indicate that the proposed regularization methods could jointly improve the performance of the baseline model.

\begin{figure}[!t]
   \subfloat[][]{
      \includegraphics[scale=0.6]{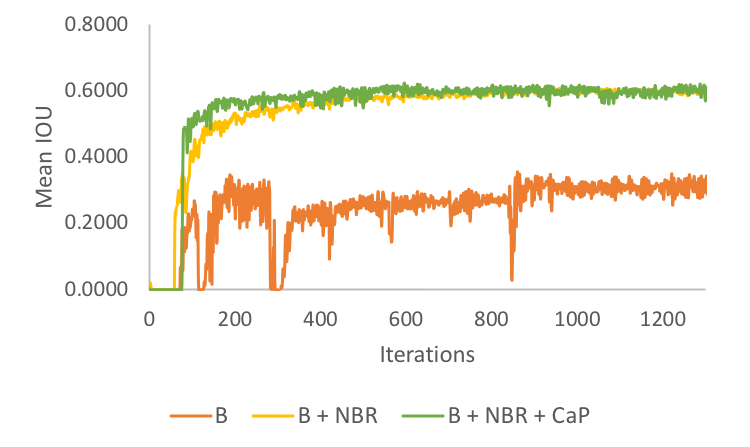}
      \centering
      \label{1_shot_conv}
   }\\
   \subfloat[][]{
      \includegraphics[scale=0.6]{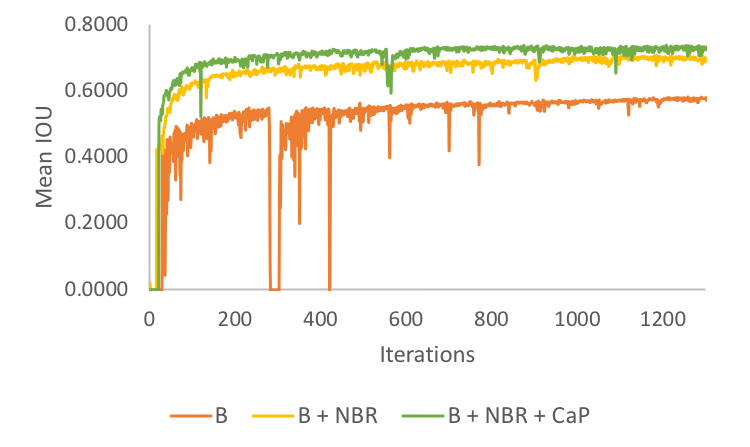}
      \centering
      \label{5_shot_conv}
   }
   \label{fig:convergence}%
   \caption{The mean IOUs using the testing set of category ``Carpet'' during the training of three ablation models under (a) 1-shot setting and (b) 5-shot setting. The iteration shown along the horizontal axis refers to the inner loop iteration shown in Algorithm 1. Best viewed in the color version.}
\end{figure}

To observe the regularization performance, we further plot the curves of mean IOUs using the testing set of category ``Carpet'' during the training of the three ablation models. Figs.~\ref{1_shot_conv} and \ref{5_shot_conv} shows the mean IOUs under the settings of 1-shot and 5-shot, respectively. From the figures, it is observed that, compared with the baseline \textbf{B}, both \textbf{B+NBR} and \textbf{B+NBR+CaP} not only achieve better generalization performance, but also accelerate the training convergence process by adding effective regularization.

\begin{table}[htpb]
  \centering
  \caption{mean IOU and DC for ablation study on the proposed regularization methods.}
    \begin{tabular}{|c|cc|cc|}
    \toprule{}
    \multirow{2}[2]{*}{Method} & \multicolumn{2}{c|}{1 shot} & \multicolumn{2}{c|}{5 shot} \\
          & IOU   & DC  & IOU   & DC \\
    \midrule
    \midrule
    B     & 0.3190 & 0.3936 & 0.5507 & 0.6537 \\
    B+NBR & 0.4241 & 0.5245 & 0.6039 & 0.7101 \\
    B+NBR+CaP & \textbf{0.4919} & \textbf{0.5944} & \textbf{0.6445} & \textbf{0.7479} \\
    \bottomrule
    \end{tabular}%
  \label{tab:ablation_IOU_DICE}%
\end{table}%

\begin{figure}[htpb]
  \centering
  \includegraphics[scale=0.3]{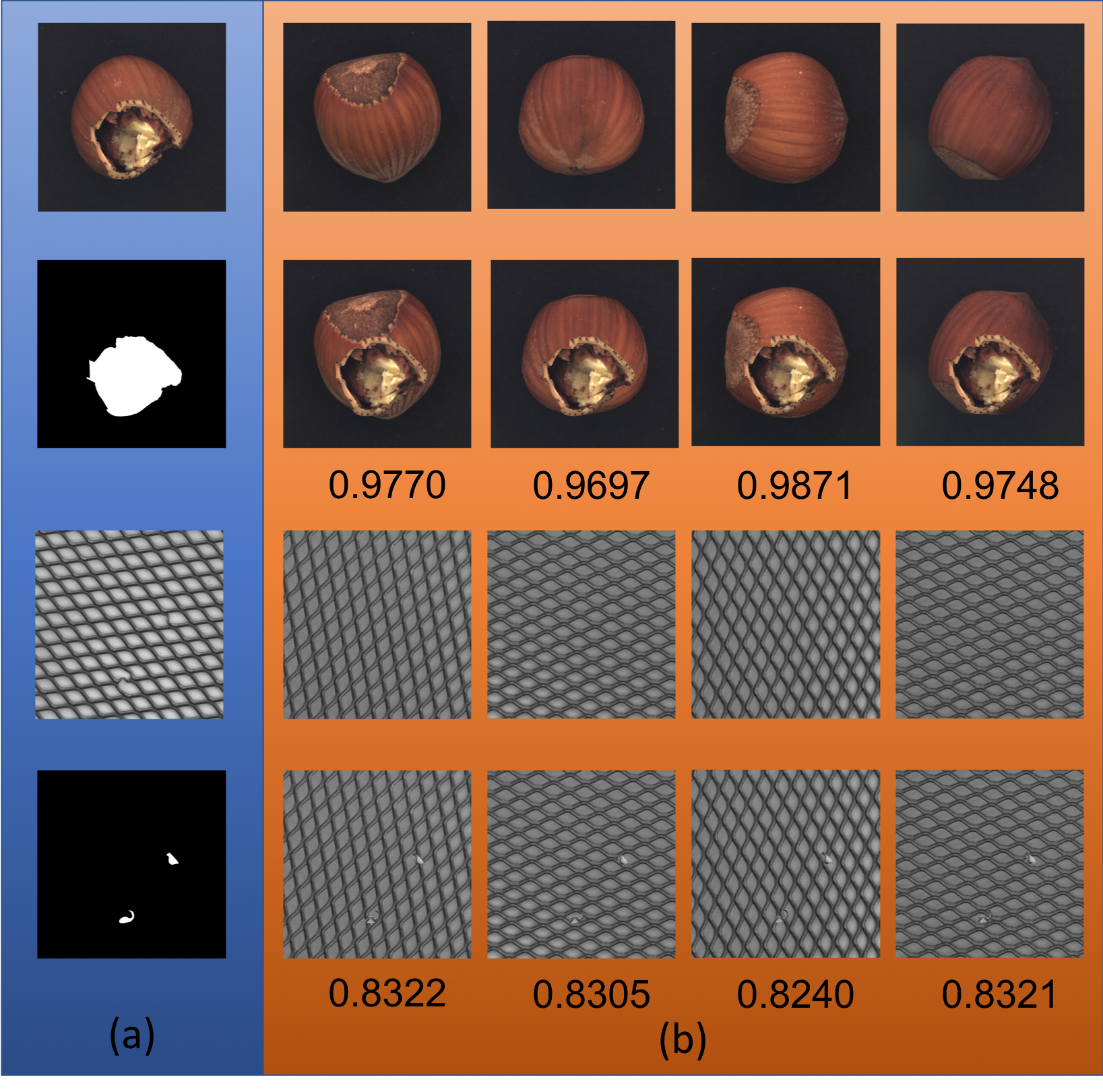}\\
  \caption{The visualization examples of category ``Hazelnut'' and ``Tile'' for the study of $\lambda$ proposed in Section~\ref{subsec:DCP}. (a) The original anomalous images and their defect groundtruth mask. (b) The normal images and the augmented images by crop-and-paste operation. The numbers below the augmented images are the values of $\lambda$.}
  \label{fig:lambda_ablation}
\end{figure}

We also study the effect of exploiting the weight $\lambda$ to quantify the ``realistic degree'' of the crop-and-pasted augmented image as proposed in Section~\ref{subsec:DCP} and Eq. (\ref{eq:lambda}). Fig.~\ref{fig:lambda_ablation} shows the visualization examples of category ``Hazelnut'' and ``Tile'' for different $\lambda$ values. During the Crop-and-Paste operation, we reserve the spatial location of the pasted defective region in the original anomalous image to avoid generating unrealistic anomalous training images. We further use the parameter $\lambda$ to finely quantify the ``realistic degree'' of the generated image. It is observed that more realistic augmented anomalous images (e.g., the four images for the ``Hazelnut'' category in Fig.~\ref{fig:lambda_ablation}) are reflected by larger $\lambda$ values ($\ge$ 0.95). The less realistic augmented anomalous images are quantified by smaller $\lambda$ values, such as the cases of ``Tile'' where the background pattern directions and illumination conditions are quite different between the original anomalous images and the selected normal images. We conduct the experiments on comparing our \textbf{B+NBR+CaP} with its counterpart by fixing the value of $\lambda$ as 1. The mean IOUs and DCs are shown in Table~\ref{tab:ablation_lambda}. By introducing the weight $\lambda$, the segmentation performance is improved for both 1-shot and 5-shot settings. It is also noted that $\lambda$ is not a weight need to be tuned but can be directly determined by Eq. (\ref{eq:lambda}) during training.

\begin{table}[htpb]
  \centering
  \caption{Mean IOU and DC for ablation study on $\lambda$}
    \begin{tabular}{|c|cc|cc|}
    \toprule
    \multirow{2}[2]{*}{Method} & \multicolumn{2}{c|}{1 shot} & \multicolumn{2}{c|}{5 shot} \\
          & IOU   & DC  & IOU   & DC \\
    \midrule
    \midrule
    B+NBR+CaP ($\lambda=1$) & 0.4745 & 0.5728 & 0.6371 & 0.7398 \\
    Ours  & \textbf{0.4919} & \textbf{0.5944} & \textbf{0.6445} & \textbf{0.7479} \\
    \bottomrule
    \end{tabular}%
  \label{tab:ablation_lambda}%
\end{table}%

For NBR loss, the cosine similarity is adopted to measure the alignment of two normal components inspired by several few-shot learning literatures~\cite{vinyals2016matching,zhang2019canet,wang2019panet}. Here, we consider the ablation model which uses Euclidean distance to replace the cosine similarity in Eq. (\ref{eq:LNBR}). Table~\ref{tab:ablation_nbr} shows that both similarity metrics could produce comparable performance under 1-shot setting while our version with cosine similarity outperforms the Euclidean distance based one by around 2.5\% in mean IOU and DC under 5-shot setting. Based on this analysis, the cosine similarity is adopted in our NBR loss.

\begin{table}[htbp]
  \centering
  \caption{Mean IOU and DC for ablation study on NBR loss}
    \begin{tabular}{|c|cc|cc|}
    \toprule
    \multirow{2}[2]{*}{Method} & \multicolumn{2}{c|}{1 shot} & \multicolumn{2}{c|}{5 shot} \\
          & IOU   & DC  & IOU   & DC \\
    \midrule
    \midrule
    B+NBR+CaP (Euclidean) & 0.4954 & 0.5947 & 0.6186 & 0.7223 \\
    B+NBR+CaP (Cosine) & 0.4919 & 0.5944 & 0.6445 & 0.7479 \\
    \bottomrule
    \end{tabular}%
  \label{tab:ablation_nbr}%
\end{table}%

For computational complexity, both NBR and CaP do not introduce any new tunable parameter or expand the baseline network during inference. Hence, the computational complexity remains the same as the baseline network while the segmentation performance is improved.



\section{Conclusion}
\label{sec:conclusion}
To handle the few-shot defect segmentation problem, we have proposed two effective regularization methods, namely Normal Background Regularization (NBR) and Crop-and-Paste (CaP) operation into the training of a U-Net-like segmentation network. Both methods effectively exploit the abundant defect-free normal images to alleviate the model overfitting due to very limited annotated anomalous training images. NBR facilitates the encoder network to produce distinctive representations for normal regions. CaP exploits normal training images for data augmentation and enhance the training of the segmentation networks by emphasizing more realistic augmented images. Extensive experiments are conducted on MVTec Anomaly Detection dataset across both object and texture category images. The experimental results show that, both quantitatively and qualitatively, the proposed methods significantly outperform multiple benchmarking methods under the settings of few-shot defect segmentation using high-resolution industrial images.

\bibliographystyle{IEEEtran}
\bibliography{OurRef}
\end{document}